\pdfoutput=1

\documentclass[11pt]{article}

\usepackage{emnlp2021}

\usepackage{times}
\usepackage{latexsym}

\usepackage[T1]{fontenc}

\usepackage[utf8]{inputenc}

\usepackage{microtype}

\usepackage{amsmath,amssymb,amsfonts}
\usepackage{booktabs}
\usepackage{graphicx}
\usepackage{subcaption}
\usepackage{algorithm} 
\usepackage{algpseudocode}
\usepackage{multirow}
\usepackage{makecell}
\usepackage{hyperref}
\usepackage{bbm}

\title{NoiER: An Approach for Training more Reliable Fine-Tuned Downstream Task Models}

\author{\textbf{Myeongjun Jang} \normalfont{and} \textbf{Thomas Lukasiewicz} \\
    University of Oxford \\
    Department of Computer Science \\
    \texttt{\{myeongjun.jang, thomas.lukasiewicz\}@cs.ox.ac.uk}}

\date{}

\begin{document}
\maketitle
\begin{abstract}
The recent development in pretrained language models trained in a self-supervised fashion, such as BERT, is driving rapid progress in the field of NLP. However, their brilliant performance is based on leveraging syntactic artifacts of the training data rather than fully understanding the intrinsic meaning of language. The excessive exploitation of spurious artifacts causes a problematic issue: The distribution collapse problem, which is the phenomenon that the model fine-tuned on downstream tasks is unable to distinguish out-of-distribution (OOD) sentences while producing a high confidence score. In this paper, we argue that distribution collapse is a prevalent issue in pretrained language models and propose \emph{noise entropy regularisation (NoiER)} as an efficient learning paradigm that solves the problem without auxiliary models and additional~data. The proposed approach improved traditional OOD detection evaluation metrics by 55\% on average compared to the original fine-tuned models.
\end{abstract}

\section{Introduction}
Contextualised representations of pretrained language models (PLMs) like ELMo \cite{ELMo}, BERT \cite{BERT}, and GPT \cite{GPT} have accomplished promising results in  many NLP tasks. 
LMs are fine-tuned on downstream tasks while achieving  state-of-the-art (SOTA) performance, even exceeding that of humans in several datasets \cite{SQUAD1.1, SWAG}. This new paradigm has become a predominant framework that paves the way for immense progress in  NLP. Many novel approaches have been proposed to obtain more elaborate contextualised representations \cite{Xlnet, Electra} or to reduce the size of the model while maintaining its performance \cite{DistilBERT}. Based on the promising results of pre-trained LMs, numerous studies argue that they are capable of \textit{understanding} human language \cite{BERT, ohsugi2019simple}.

However, recent studies reveal that the performance of PLMs does not arise from human-level language understanding but is based on their ability to excessively leverage syntactic patterns of training data. \citet{niven2019probing} claim that BERT’s performance is entirely based on the exploitation of spurious statistical cues that are present in the training set. They modified the English Argument Reasoning Comprehension dataset \cite{habernal2018argument} by adding adversarial examples that just one token of the original was negated. BERT’s performance on the modified dataset falls short of expectations as a consequence of the excessive usage of the cues in the original training set. Similarly, \citet{mccoy2019right} figure out that BERT is only highly competent in leveraging syntactic heuristics. Experiments on the novel HANS dataset, which \citet{mccoy2019right} designed to frustrate the heuristics, exhibits a significant decrease in BERT’s performance. \citet{benderkoller2020climbing} argue that the widespread belief that pre-trained LMs \textit{understand} or \textit{comprehend} is a hype. Based on the result of carefully designed thought experiments, \citet{benderkoller2020climbing} show that it is impossible to learn the meaning of language by only leveraging the \textit{form} of sentences.

The model that lacks generalisation ability tend to fail on data from a different distribution, providing high-confidence predictions while being incorrect \cite{goodfellow2015explaining, amodei2016concrete, hendrycks2017baseline}. Therefore, the results of the above studies cast the following question: 

\medskip 
\textit{Can PLMs distinguish in-distribution (IND) and out-of-distribution (OOD) sentences when fine-tuned on downstream tasks?} 

\medskip 
In the \textit{open} real world, there exist a huge number of OOD instances. Therefore, even though the models perform brilliantly on a closed-distribution dataset, there is a low chance for the models to be employed in practice, provided that they are incapable of rejecting OOD sentences. Generating incorrect results on these inputs and conducting unwanted commands may result in disasters, which undermines the reliability and reputation of a model. It could also trigger even more severe outcomes, particularly in risk-sensitive applications. Hence, to insist that a model is intelligent and can truly understand language, it should at least be capable of rejecting OOD sentences, just as humans~do.

In this paper, we empirically show that distribution collapse, the phenomenon that a model yields high confidence scores on OOD instances and fails to distinguish them from IND instances, is a prevalent issue in the pretraining and fine-tuning framework, even for a simple text classification task. Next, we propose a simple but effective approach, called \emph{noise entropy regularisation (NoiER)}, which has the ability to reject OOD sentences  while maintaining a high classification performance. We experimentally confirm that our proposed approach not only alleviates distribution collapse but also outperforms baseline models devised for detecting OOD sentences. The main contributions of this paper can be briefly summarised as follows:

\begin{itemize}
    \item Through experiments, we reveal that large pre-trained LMs suffer from the distribution collapse issue when fine-tuned on downstream tasks.
    \item We propose noise entropy regularisation (Noi\-ER) as an efficient method that alleviates the distribution collapse problem without leveraging additional data or adding auxiliary models.
    \item Compared to the fine-tuned models, our approach produces better AUROC and EER, which are increased by 60\% and 49\%, respectively, while maintaining a high classification capacity.
\end{itemize}

\section{Related Works}\label{related_works}
OOD detection plays a crucial role in industrial processes for ensuring reliability and safety \cite{HARROU2016365}. As a result, numerous studies have been made in various fields, including NLP. 

\noindent \textbf{Classification.} A simple way to classify OOD instances is to add the ``OOD'' class to a classifier \citet{kim2018joint}. However, this approach is practically inefficient, considering the enormously broad distribution of OOD data. Another approach is detecting OOD sentences in an unsupervised manner by using IND samples only, which is known under several different names, such as \textit{one-class classification} or \textit{novelty detection}. Some works that leverage distance-based \cite{8367203, xu2019open} or density-based \cite{SONG201947, SEO2020113111} methods have been proposed for text OOD detection. However, most of these approaches are computationally expensive for both training and inference, which interrupts their usage in practical applications.

\noindent \textbf{GAN-based approach.} \citet{ryu2018out} leveraged a generative adversarial network (GAN) for OOD sentence detection. They expected the discriminator to reject OOD sentences by training it to distinguish IND sentences from fake instances generated by the generator. \citet{ryu2018out} used pretrained sentence representations for training the GAN.

\noindent \textbf{Threshold-based approach.} This approach composes an OOD detector $g(x)$ that judges an instance is OOD if the criterion score is lower than a threshold, i.e., $g(x)=\mathbbm{1}\,(S(x)>\alpha)$, where $\alpha$ denotes a threshold, and $S(x)$ refers to the predicted criterion score. \citet{hendrycks2017baseline} leveraged the maximum softmax probability (MSP) as the criterion. \citet{ryu2017neural} trained a text classifier and an autoencoder, and used the reconstruction errors for discerning OOD sentences.

\citet{liang2018enhancing} proposed a method, called ODIN, which improved the  MSP-based approach \cite{hendrycks2017baseline} by applying temperature scaling and input perturbations. Temperature scaling modifies the softmax output as follows:
\begin{equation}
\begin{gathered}
S_i(\textrm{x};T) = \frac{exp(f_i(\textrm{x})/T)}{\sum_{i=1}^{N} exp(f_i(\textrm{x})/T)},
\end{gathered}
\end{equation}
where $T \in \mathbb{R}^+$ denotes the temperature scaling parameter and is set to 1 during  training. Next, during the inference process, a small perturbation is added to the input as follows:
\begin{equation}
\begin{gathered}
\Tilde{\textrm{x}}=\textrm{x}-\epsilon sign(-\nabla_x log(S_{\hat{y}}(\textrm{x};T))),
\end{gathered}
\end{equation}
where $\epsilon$ is the perturbation magnitude and $S_{\hat{y}}(\textrm{x};T)$ $=$ $\max_{i} S_i(\textrm{x};T)$. Although ODIN is designed for image classification,  it is applicable to any type of classification models. This method has a practical benefit in that model training is unnecessary.

\noindent \textbf{Entropy regularisation.} The entropy regularisation is designed to make a threshold-based approach better detect OOD. The model is trained to minimise both cross-entropy loss ($\mathcal{L}_{ce}$) and entropy regularisation loss ($\mathcal{L}_{er}$) that maximises the entropy of the predictive distribution for OOD instances, while making it closer to the uniform distribution. This leads the model to conduct correct predictions for IND samples, while being less confident on OOD instances, and hence improves the performance of the predictive distribution threshold-based methods.

In computer vision, \citet{lee2018training} trained a GAN-based OOD generator to produce OOD examples and used them for minimising $\mathcal{L}_{er}$. \citet{hendrycks2018deep} leveraged out-of-distribution datasets to maximise the entropy. \citet{zheng2020out} proposed a model called ER-POG to detect OOD in dialog systems. They observed that generating OOD samples in the continuous space \cite{ryu2018out, lee2018training} is inefficient for text data. Instead, they leveraged an LSTM-based sequence-to-sequence (Seq2Seq) autoencoder and GAN to generate discrete token OOD samples.

\section{Distribution Collapse}
In this section, we delve into the distribution collapse issue of PLMs. We experimentally  verify that the distribution collapse truly exists in PLMs when they are fine-tuned on downstream tasks.

\subsection{IND and OOD Disparity}
Assume that a text classification model $M$ is trained to classify a category $C_i \in \{C_1,\ldots, C_K\}$ that belongs to IND.The model can detect OOD by rejecting examples that generate threshold-below MSP scores  \cite{hendrycks2017baseline}. Therefore, it should produce a high MSP score for an IND instance to not reject it. On the contrary, the predictive distribution for an OOD instance should be close to the uniform distribution, i.e., zero confidence \cite{lee2018training, hendrycks2018deep, zheng2020out}. Therefore, we first define a uniform disparity (UD) for a given sentence set $S$ as follows:
\begin{equation}
\begin{gathered}
UD(S, M) = \frac{1}{|S|} \sum_{s_i \in S} JSD(M(s_i), u),
\end{gathered}
\end{equation}
where $u$ refers to the uniform distribution, and $JSD$ denotes the Jensen-Shannon divergence. Next, for a given IND set $I$ and OOD set $O$, the IND-OOD disparity (IOD) is calculated as follows:
\begin{equation}
\begin{gathered}
\!\!\!\!IOD(I,O,M) = UD(I,M) - UD(O,M).
\end{gathered}
\end{equation}
Note that the metric is higher the better. Also, it can generate a negative value, which is the most undesirable case.

\begin{table}[t!]
	\begin{center}
		\caption{Statistics of datasets for the distribution collapse experiment} \label{table1.dataset}%
		\renewcommand{\arraystretch}{1.3}
		\footnotesize{
			\centering{\setlength\tabcolsep{3pt}
				\begin{tabular}{cccc}
					\toprule
					\hline
					& AG's News & Fake News & Corona Tweets \\
					\hline
					\# of classes & 4 & 2 & 5 \\
				    Training set size & 120,000 & 40,408 & 41,157 \\
					Test set size & 7,600 & 4,490 & 3,798 \\
                    \makecell{Avg sentence \\ length} & \makecell{9.61 \\ tokens} & \makecell{17.02 \\ tokens} & \makecell{41.85 \\ tokens} \\
					\hline
					\bottomrule	
		\end{tabular}}}
	\end{center}
\end{table}

\subsection{Experiments}\label{section3-2}
\textbf{Datasets.} For the experiments, we collected three datasets from different domains and tasks: AG's News dataset \cite{AGNews} for topic classification, the Fake News classification dataset \cite{FakeNews}, and the Corona Tweets dataset\footnote{\href{https://www.kaggle.com/datatattle/covid-19-nlp-text-classification}{https://www.kaggle.com/datatattle/covid-19-nlp-text-classification}}, which is designed to classify the sentiment of tweets regarding COVID 19. For AG's News and Fake News data, we used article titles as input text. We only conducteded minor preprocessing, such as removing the URL and special symbols for a hashtag. The statistics of each dataset are presented in Table \ref{table1.dataset}.

\begin{table*}[t!]
	\begin{center}
		\caption{Results for the verification of distribution collapse. The name of the dataset corresponds to the IND data. We trained each model 10 times and recorded the average of each metric. The notation ``$\uparrow$'' denotes that higher values are better, and ``$\downarrow$'' stands for lower values are better. For the IOD value, we multiplied the original figure with 100 unify the scale with the other metrics. The best values are formatted in bold. The figures of fine-tuned models show a significant difference compared to TFIDF-NB with $p$-value $<$ 0.05 ($\dagger$) and $p$-value $<$ 0.01 (*) using the t-test.} \label{table2.Original_result}%
		\renewcommand{\arraystretch}{1.3}
		\footnotesize{
			\centering{\setlength\tabcolsep{3.5pt}
		\begin{tabular}{l|llll|llll|llll}
		\toprule
		\hline
		\multirow{2}{*}{Model} & \multicolumn{4}{c|}{AG's News} & \multicolumn{4}{c|}{Fake News} & \multicolumn{4}{c}{Corona Tweets}\\ 
		& F1$\uparrow$ & IOD$\uparrow$ & AUROC$\uparrow$ & EER$\downarrow$
		& F1$\uparrow$ & IOD$\uparrow$ & AUROC$\uparrow$ & EER$\downarrow$ 
		& F1$\uparrow$ & IOD$\uparrow$ & AUROC$\uparrow$ & EER$\downarrow$ \\ \hline	
        BERT & 89.40* & 7.25* & 75.87* & 31.20*
        & 98.55* & 0.64$\dagger$ & 58.06 & 44.89 & \textbf{83.17}* & -2.13* & 40.65* & 57.51* \\
        DistilBERT & \textbf{89.82}* & 7.35* & 76.96* & 30.48*
        & \textbf{98.62}* & 1.48 & \textbf{59.94}$\dagger$ & \textbf{44.64} & 81.61* & -2.24* & 41.63* & 57.02* \\
        Electra & 88.91* & 7.12* & 79.05* & 28.11*
        & 98.48* & \textbf{1.91} & 57.14 & 47.79 & 82.05* & -1.85* & 42.28* & 56.61* \\
        SqueezeBERT & 89.22* & 6.05* & 74.96* & 32.27*
        & 98.46* & 1.75 & 55.45 & 48.89 & 82.38* & -1.83* & 41.94* & 56.83* \\
        ERNIE 2.0 & 89.45* & 5.85* & 73.78* & 33.07*
        & 98.56* & 1.82 & 50.72 & 53.06 & 82.12* & -2.13* & 41.00* & 56.75* \\ \hline
        TFIDF-NB & 86.96 & \textbf{14.94} & \textbf{83.10} & \textbf{23.67}
        & 95.93 & 1.35 & 52.09 & 49.34 & 37.10 & \textbf{5.99} & \textbf{73.20} & \textbf{33.19}\\ \hline
		\bottomrule
		\end{tabular}}}
	\end{center}
\end{table*}

\noindent \textbf{Experimental Design.} To minimise the overlap between each dataset, we conducted experiments on the following three IND-OOD pairs: (1) AG's News -- Corona Tweets, (2) Fake News -- Corona Tweets, and (3) Corona Tweets -- AG's News and Fake News. We first trained an IND classifier and measured the F1-score. Next, OOD examples are used to calculate the IOD score. We also measured AUROC and the equal error rate (EER) for evaluating the OOD detection performance. An illustration of these metrics are provided in Appendix~\ref{appendix.eval_metric}. Following the work of \citet{hendrycks2017baseline}, we used the MSP as confidence score.

\noindent \textbf{Model Candidates.} We fine-tuned five PLMs for the experiments: BERT \cite{BERT}, DistilBERT \cite{DistilBERT}, ELECTRA \cite{Electra}, SqueezeBERT \cite{SqueezeBERT}, and ERNIE 2.0 \cite{ERNIE2}. We used the  transformers package\footnote{\href{https://github.com/huggingface/transformers}{https://github.com/huggingface/transformers}} provided by Hugging Face. To avert overfitting as much as possible, we applied dropout \cite{Dropout}. In addition, we trained a simple TF-IDF-based na\"ive Bayes classifier (TFIDF-NB) as baseline.

\noindent \textbf{Training Details.} During  fine-tuning, 10\% of the training set is used as  validation set for early stopping. We used the AdamW optimiser \cite{AdamW} for training. The learning rate was 1e-6 for ERNIE 2.0, and 1e-4 for the others.\footnote{We used a single GeForce GTX TITAN XP GPU for training the models.}

\begin{figure}[t!]
	\centering
	\begin{subfigure}[b]{0.4\textwidth}
		\includegraphics[width=\linewidth]{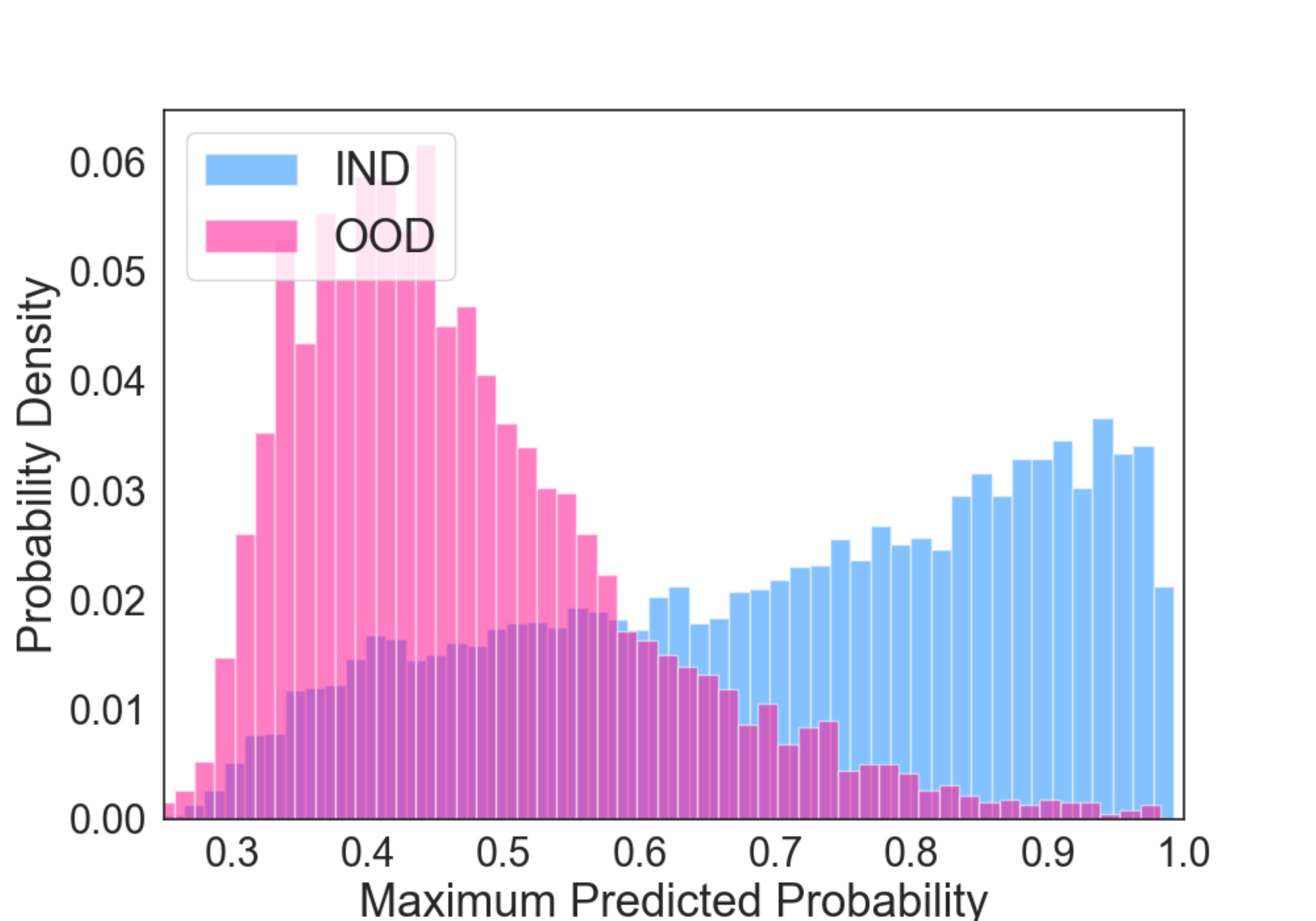}
		\caption{TFIDF-NB}
	\end{subfigure} \
	\begin{subfigure}[b]{0.4\textwidth}
		\includegraphics[width=\linewidth]{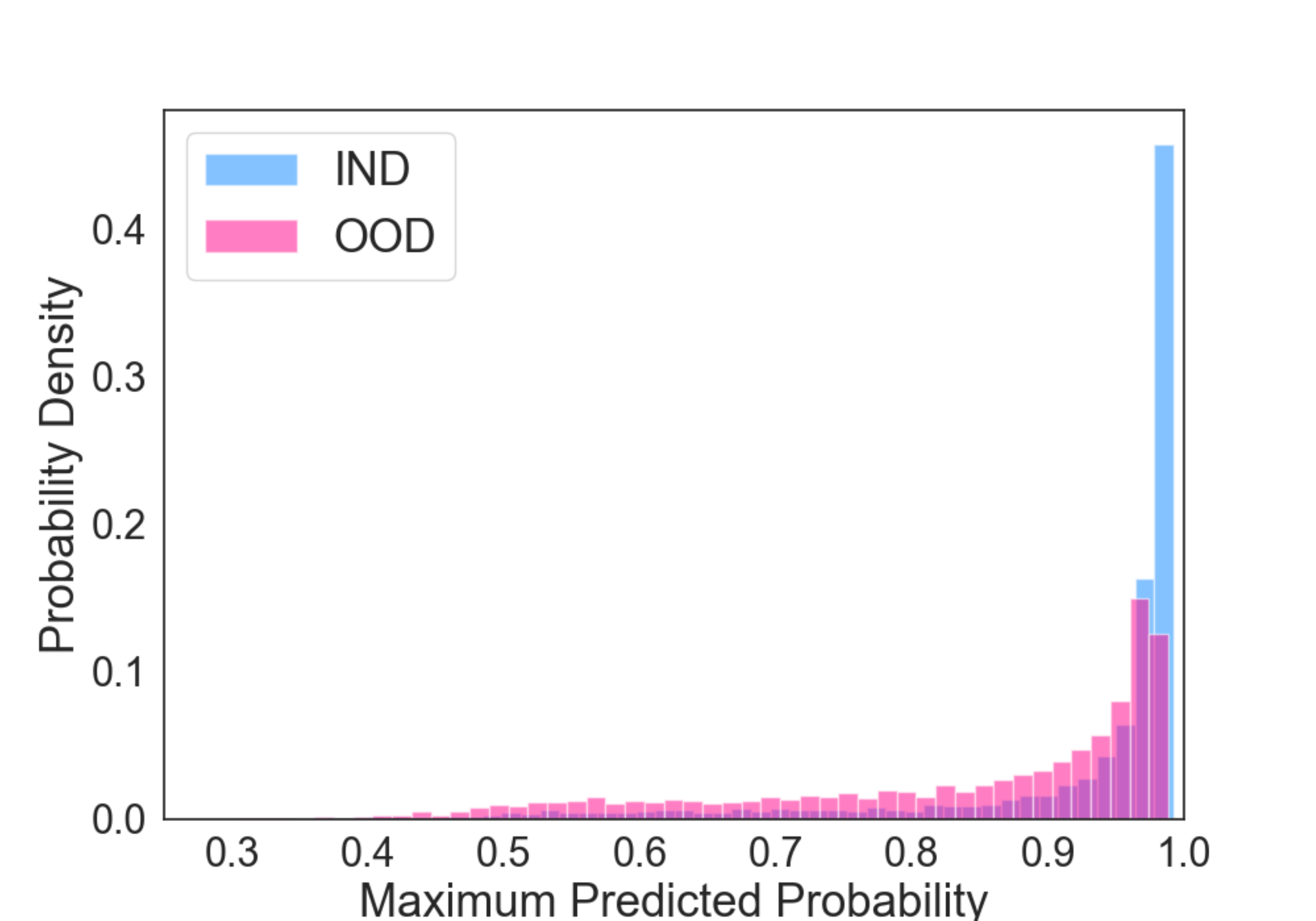}
		\caption{BERT-base}
	\end{subfigure}%
	\caption{Maximum predicted probability distribution plots when the AG's News dataset is IND.}
	\label{fig:dist_msp}
\end{figure}

\noindent \textbf{Results.} The experimental results are shown in Table \ref{table2.Original_result}. Although the fine-tuned PLMs outperformed TFIDF-NB in terms of classification ability, their OOD detection performance fell short of expectation. In particular, they performed less than TFIDF-NB in the AG's News dataset and entirely failed to detect the OOD in the Corona Tweets dataset while generating negative IOD values. Although the PLMs exhibited better results compared to TFIDF-NB in the Fake News dataset, the difference was not statistically significant apart from DistilBERT. Even ERNIE 2.0, which leverages both text form and external knowledge, failed to yield a promising result. Figure \ref{fig:dist_msp} depicts the distributions of the maximum predicted probability of  TFIDF-NB and BERT-base on the AG's News dataset. BERT-base produces very high confidence scores on OOD examples contrary to TFIDF-NB, which is capable of generating a low confidence score on OOD instances. Likewise, the experimental results reveal that PLMs are not competent in detecting OOD sentences when they are fine-tuned in downstream tasks. This supports our claim that PLMs suffer from the distribution collapse issue, which harms the model's reliability. 

\section{NoiER: Noise Entropy Regularisation}\label{noier}
The most suspicious cause of distribution collapse is the excessive exploitation of syntactic artifacts \cite{niven2019probing, mccoy2019right}. Therefore, improving the model's generalisation ability by preventing the model from learning spurious artifacts is essential to overcome this issue. 
To this end, we applied the entropy regularisation (ER) technique, which has shown promising performance in improving the model's robustness \cite{hendrycks2018deep, lee2018training, zheng2020out}. The question is which sentences could be leveraged for minimising the ER term. Collecting OOD instances is practically inefficient, as we illustrated in Section \ref{related_works}. Training an OOD sentence generator \cite{lee2018training, zheng2020out} also has the demerit that the model becomes highly contingent on the generator's performance. Also, we have to retrain both the OOD sample generator and classifier whenever the distribution of training data changes, which is a time- and resource-intensive~work. 

To overcome the practical disadvantage, we decided to leverage a model-free approach by considering noise-added IND sentences as OOD data points. The role of the perturbed sentences is to assist the model with recognising the fact that a data point could be included in OOD even though it has analogous syntactic patterns with IND sentences. Hence, they should satisfy the following conditions:

\smallskip 
\noindent \textbf{Condition 1.} A noise sentence should share a similar form with IND sentences to a certain extent.

\smallskip 
\noindent \textbf{Condition 2.} Despite Condition 1, a noise sentence must be considered as OOD sentence.

\subsection{Noise Generation Scheme}
We used three noise generation functions: deletion, permutation \cite{lample2017unsupervised}, and replacement.

\begin{enumerate}
    \item \textbf{Deletion}: This function removes words in a sentence $s$ with the probability of $p_{del}$.
    \item \textbf{Permutation}: This function shuffles at most $r_{perm}\%$ of words in a sentence $s$. 
    \item \textbf{Replacement}: This function replaces words in a sentence $s$ to randomly generated words with the probability of $p_{repl}$.
\end{enumerate}

We distinguished the terminology ``word'' from ``token'',  because our approach focuses on generating noise on the word level rather than subword (i.e.,  token) level. Note that the noise generation methods are independent from the model's tokenizer. Each noise function has its corresponding hyperparameters. We empirically found the optimum values for each dataset while analysing the IOD value of the validation dataset. More detailed explanations regarding the hyperparameter search are given in Appendix~\ref{appendix.hps}.

\begin{algorithm}
\caption{Noise entropy regularisation training process for fine-tuning model $M_\theta$.}
	
\textbf{Input:} An input sentence set $X=(x_1, x_2, ...,x_n)$,  corresponding label set $Y=(y_1, y_2, ..., y_n)$, number of iterations $T$.\\
\textbf{Output:} A fine-tuned model $M_{\theta}$
\begin{algorithmic}[1]
\For{iter = 1, ..., T}
	\State $X_b, Y_b$ $\leftarrow$ \textrm{BATCH(X,Y)};
	\State ${\Tilde{X_b}}$  $\leftarrow$ \textrm{NOISE($X_b$)};
	\State $\hat{Y}$ $\leftarrow$ $M_{\theta}(X_b)$;
	\State $\mathcal{L}_{CE}$ $\leftarrow$ $\frac{1}{|X_b|}\sum_{y_i \in \hat{Y}} -y_i \log(\hat{y}_i)$;
	\State $\mathcal{L}_{ER}$ $\leftarrow$ $\frac{1}{|\Tilde{X_b}|} \sum_{\Tilde{{x_i}} \in \Tilde{X_b}} JSD(M_{\theta}(\Tilde{x_i}), u)$;
	\State $\mathcal{L} \leftarrow \mathcal{L}_{CE} + \mathcal{L}_{ER}$;
	\State Update $\theta$ by descending its gradient $\nabla_{\theta}\mathcal{L}$
	\EndFor
\end{algorithmic}
\label{algo.training}
\end{algorithm} 

\subsection{Model Training}
For each training iteration, we generated a noise sentence set $\Tilde{X}_b$. Next, we leveraged $\Tilde{X}_b$ to reduce the ER term. For a given model $M_{\theta}$, we defined the ER loss as the Jensen-Shannon divergence between the predicted probabilities and uniform distribution to reflect minimising the uniform disparity directly to our training objective:
\begin{equation}
\begin{gathered}
\mathcal{L}_{ER}(\theta) = \sum_{\Tilde{x_i} \in \Tilde{X_b}} JSD(M_{\theta}(\Tilde{x_i}),u),
\end{gathered}
\end{equation}
where $JSD$ and $u$ denote the Jensen-Shannon divergence and the uniform distribution, respectively. Finally, the final training loss is defined as the sum of the cross-entropy loss and the ER loss:
\begin{equation}
\begin{gathered}
\mathcal{L}(\theta)=\mathcal{L}_{CE}(\theta) + \alpha\mathcal{L}_{ER}(\theta),
\end{gathered}
\end{equation}
where $\alpha$ is a hyperparameter, which is set to $\alpha=1$.\footnote{We confirmed that the results are not sensible to $\alpha$.} The overall training process is shown in Algorithm \ref{algo.training}.

\begin{table*}[t!]
	\begin{center}
		\caption{Results for the ablation study on DistilBert. The name of the dataset corresponds to IND. We trained each model 10 times and recorded the average of each metric. The best values are in bold. The notation ``$\uparrow$'' denotes that higher values are better, and ``$\downarrow$'' stands for lower values are better. For the IOD value, we multiplied the original figure with 100  to unify the scale with the other metrics. ``Full'' refers to leveraging the whole noise functions, whereas ``w/o'' denotes removing the following noise function.
		} \label{table3.AblationStudy}%
		\renewcommand{\arraystretch}{1.3}
		\footnotesize{
			\centering{\setlength\tabcolsep{2.5pt}
		\begin{tabular}{l|cccc|cccc|cccc}
		\toprule
		\hline
		\multirow{2}{*}{Model} & \multicolumn{4}{c|}{AG's News} & \multicolumn{4}{c|}{Fake News} & \multicolumn{4}{c}{Corona Tweets}\\ 
		& F1$\uparrow$ & IOD$\uparrow$ & AUROC$\uparrow$ & EER$\downarrow$ 
		& F1$\uparrow$ & IOD$\uparrow$ & AUROC$\uparrow$ & EER$\downarrow$
		& F1$\uparrow$ & IOD$\uparrow$ & AUROC$\uparrow$ & EER$\downarrow$ \\ \hline	
        Full & 93.46 & 23.54 & 87.57 & 19.40 & 98.92 & 13.90 & 83.51 & \textbf{24.19} & 91.24 & 36.97 & 92.43 & 14.54 \\ 
        w/o Deletion & 93.44 & 35.02 & 91.47 & 16.08 & \textbf{99.35} & 20.22 & 81.55 & 28.66 & 91.26 & 18.24 & 79.17 & 28.59 \\ 
        w/o Permutation & 92.99 & 17.16 & 83.63 & 23.72 & 99.02 & 11.67 & 82.58 & 25.32 & 90.58 & 38.07 & 93.41 & 13.41 \\
        w/o Replacement & 92.45 & 13.54 & 76.27 & 28.82 & 98.85 & 9.54 & 75.99 & 31.55 &	91.94 & 40.09 & 93.71 & 12.69 \\
        only Deletion & 91.98& 5.48 & 64.55 & 38.44 & 98.77 & 6.33 & 71.75 & 34.52 & 90.58 & \textbf{41.28} & \textbf{94.96} & \textbf{10.69} \\
        only Permutation & 93.22 & 14.86 & 76.33 & 29.75 & 98.83 & 7.53 & 59.35 & 46.53 & 92.32 & 20.76 & 78.77 & 28.54\\ 
        only Replacement & \textbf{94.13} & \textbf{38.37} & \textbf{93.57} & \textbf{13.97} & 99.17 & \textbf{20.28} & \textbf{83.81} & 27.97 & \textbf{92.86} & 8.85 & 70.60 & 35.86 \\ \hline
		\bottomrule
		\end{tabular}}}
	\end{center}
\end{table*}

\begin{table*}[t!]
	\begin{center}

		\caption{Results for the comparison with the original fine-tuning framework. The name of the dataset corresponds to IND. We trained each model 10 times and recorded the average of each metric. For each pre-trained models, the best evaluation metric values are in bold. The notation ``$\uparrow$'' denotes that higher values are better, and ``$\downarrow$'' stands for lower values are better. ``Full'' refers to using all noise functions, whereas ``Best'' denotes the noise function combination that showed the best OOD detection performance. For the IOD value, we multiplied the original figure with 100 to unify the scale with the other metrics. Our proposed approach outperforms original fine-tuned models with $p$-value $<$ 0.05 ($\dagger$) and $p$-value $<$ 0.01 (*) using the t-test.} \label{table4.Original_compare}%
		\renewcommand{\arraystretch}{1.3}
		\footnotesize{
			\centering{\setlength\tabcolsep{1.3pt}
		\begin{tabular}{cl|cccc|cccc|cccc}
		\toprule
		\hline
		\multicolumn{2}{c|}{\multirow{2}{*}{Model}} & \multicolumn{4}{c|}{AG's News} & \multicolumn{4}{c|}{Fake News} &  \multicolumn{4}{c}{Corona Tweets}\\ 
		& & F1$\uparrow$ & IOD$\uparrow$ & AUROC$\uparrow$ & EER$\downarrow$ 
		& F1$\uparrow$ & IOD$\uparrow$ & AUROC$\uparrow$ & EER$\downarrow$
		& F1$\uparrow$ & IOD$\uparrow$ & AUROC$\uparrow$ & EER$\downarrow$\\ \hline
		
        \multirow{3}{*}{\makecell{BERT}} 
        & Original & \textbf{89.40} & 7.25 & 76.96 & 30.48 & \textbf{98.55} & 0.64 & 58.06 & 44.89 & \textbf{83.17} & -2.13 & 40.65 & 57.51\\ 
        & Ours-Full & 88.39* & 18.68* & 80.60$\dagger$ & 27.16$\dagger$ & 97.81* & 13.52* & 80.44* & \textbf{27.44}* & 81.70* & 30.33* & 86.52* & 22.64* \\ 
        & Ours-Best & 89.08 & \textbf{33.26}* & \textbf{91.62}* & \textbf{16.94}* & 98.08* & \textbf{17.52}* & \textbf{82.39}* & 29.38* & 80.77* & \textbf{35.95}* & \textbf{90.82}* & \textbf{16.80}\\ \hline
        
        \multirow{3}{*}{\makecell{DistilBERT}} 
        & Original & \textbf{89.82} & 7.35 & 75.87 & 31.20 & \textbf{98.62} & 1.48 & 59.94 & 44.64 & \textbf{81.61} & -2.24 & 41.63 & 57.02\\ 
        & Ours-Full & 88.52* & 18.75* & 82.08* & 25.12* & 97.89* & 10.34* & 76.56* & \textbf{30.39}* & 80.72 & 33.15* & 88.85* & 19.25* \\ 
        & Ours-Best & 88.96* &  \textbf{37.52}* & \textbf{92.87}* & \textbf{14.24}* & 98.02* & \textbf{17.20}* & \textbf{80.19}* & 30.55* & 80.23$\dagger$ & \textbf{37.48}* &  \textbf{92.58}* & \textbf{13.78}* \\ \hline
        
        \multirow{3}{*}{\makecell{Electra}} 
        & Original & \textbf{88.91} & 7.12 & 79.05 & 28.11 & \textbf{98.48} & 1.91 & 57.14 & 47.79 & \textbf{82.05} & -1.85 & 42.28 & 56.61\\ 
        & Ours-Full & 88.11* & 17.24* & 79.85 & 27.58 & 97.93* & 10.12* & 76.09* & 32.89* & 81.34 & 30.73* & 86.38* & 22.26* \\ 
        & Ours-Best & 88.61 & \textbf{35.74}* & \textbf{91.97}* & \textbf{15.65}* & 98.16$\dagger$ & \textbf{15.40}* & \textbf{79.77}* & \textbf{32.52}* & 81.49 &	\textbf{34.39}* &	\textbf{89.24}* & \textbf{19.01}*   \\ \hline

        \multirow{3}{*}{\makecell{SqueezeBERT}} 
        & Original & \textbf{88.93} & 6.68 & 76.65 & 30.53 & \textbf{98.33} & 1.96 & 59.94 & 45.06 & \textbf{82.01} & -1.85 & 41.87 & 56.43\\ 
        & Ours-Full & 88.16* & 16.34* & 79.25$\dagger$ & 27.98$\dagger$ & 97.78$\dagger$ & 9.83* & 75.84* & \textbf{32.42}* & 81.52 & 27.29* & 83.73* & 25.25* \\ 
        & Ours-Best & 88.67* & \textbf{35.73}* & \textbf{91.67}* & \textbf{16.26}* & 98.30 & \textbf{13.78}* & \textbf{76.55}* & 35.01* & 80.76* & \textbf{36.16}* & \textbf{90.80}* & \textbf{17.16}* \\ \hline

        \multirow{3}{*}{\makecell{ERNIE 2.0}} 
        & Original & \textbf{89.45} & 5.85 & 73.78 & 33.07 & \textbf{98.56} & 1.82 & 50.72 & 53.06 & \textbf{82.12} & -2.13 & 41.00 & 56.75\\ 
        & Ours-Full & 87.96* & 18.54* & 81.24* & 26.19* & 97.95* & 12.71* & 78.24* & \textbf{28.93}* & 77.85* & 28.52* & 86.85* & 21.41* \\ 
        & Ours-Best & 88.46* & \textbf{33.77}* & \textbf{91.47}* & \textbf{16.28}* & 97.95* & \textbf{15.89}* & \textbf{79.32}* & 31.98* & 80.33* & \textbf{35.97}* & \textbf{91.21}* & \textbf{16.32}* \\ \hline
		\bottomrule
		\end{tabular}}}
	\end{center}
\end{table*}
\section{Experiments}
We conducted experiments to confirm the effectiveness of the proposed approach under the same experimental design illustrated in Section \ref{section3-2}.
\begin{table*}[t!]
	\begin{center}
		\caption{Results for the comparison with baseline models. The name of the dataset corresponds to IND. We trained each model 10 times and recorded the average of each metric. The best value is  in bold. NoiER-DistilBERT-Best showed a statistically significant difference compared to other models with $p$-value $<$ 0.05 ($\dagger$) and $p$-value $<$ 0.01 (*) using the t-test. The notation $\uparrow$ denotes that higher values are better, and $\downarrow$ stands for lower values are better. For the IOD value, we multiplied  the original figure with 100 to unify the scale with the  other metrics.} \label{table5.Benchmark_test}%
		\renewcommand{\arraystretch}{1.3}
		\footnotesize{
			\centering{\setlength\tabcolsep{1.5pt}
		\begin{tabular}{l|cccc|cccc|cccc}
		\toprule
		\hline
		\multirow{2}{*}{Model} & \multicolumn{4}{c|}{AG's News} & \multicolumn{4}{c|}{Fake News} & \multicolumn{4}{c}{Corona Tweets}\\ 
		& F1$\uparrow$ & IOD$\uparrow$ & AUROC$\uparrow$ & EER$\downarrow$ & F1$\uparrow$ & IOD$\uparrow$ & AUROC$\uparrow$ & EER$\downarrow$ & F1$\uparrow$ & IOD$\uparrow$ & AUROC$\uparrow$ & EER$\downarrow$\\ \hline

        MSP-Linear & 76.75* & 7.61* & 70.33* & 35.29* & 95.07* & 5.55* & 62.29* & 41.73* & 45.58* & 2.04* & 52.76* & 48.14* \\

        MSP-DistilBERT & 89.82* & 7.25* & 76.96* & 30.48* & 98.62* & 1.48* & 59.94* & 44.64* & \textbf{81.61}$\dagger$ & -2.24* & 41.63* & 57.02* \\
        
        ODIN-Linear & 76.75* & 7.63* & 70.38* & 35.29* & 95.07* & 5.55* & 62.30* & 41.73* & 45.54* & 2.05* & 52.78* & 48.14* \\

        ODIN-DistilBERT & \textbf{89.83}* & 7.25* & 76.97* & 30.47* & \textbf{98.63}* & 1.48* & 59.97* & 44.62* & \textbf{81.61}$\dagger$ & -2.24* & 41.66* & 57.02* \\

        AE-BiLSTMSA & 87.11* & \makecell[c]{-} & 83.74* & 24.38* & 97.90 & \makecell[c]{-} & 52.22* & 49.65* & 74.31* & \makecell[c]{-} & 68.65* & 35.97* \\

        AE-DistilBERT & 89.82* & \makecell[c]{-} & 77.37* & 28.97* & 98.62* & \makecell[c]{-} & 57.60* & 45.35* & \textbf{81.61}$\dagger$ & \makecell[c]{-} & 42.51* & 55.22* \\

        ERPOG-DistilBERT & 85.11* & 22.66* & 84.66* & 22.98* & 94.10* & -5.11* & 38.22* & 63.08* & 29.29* & -5.02* & 26.80* & 71.53* \\

        GER-DistilBERT & 89.51* & 0.73* & 72.26* & 32.87* & 98.32$\dagger$ & 0.29* & 59.60* & 46.49* & 81.34$\dagger$ & -0.50* & 44.30* & 54.86* \\
        
        CNER-DistilBERT & 89.09 & 3.73* & 71.39* & 35.37* & 97.30* & 1.81* & 46.59* & 55.05* & 78.72 & -2.95* & 40.13* & 58.16* \\
        \hline
        
        NoiER-DistilBERT-Full & 88.52$\dagger$ & 18.75* & 82.08* & 25.12* & 97.89 & 10.34* & 76.56 & \textbf{30.39} & 80.72 & 33.15* & 88.85* & 19.25* \\

        NoiER-DistilBERT-Best & 88.96 & \textbf{37.52} & \textbf{92.87} & \textbf{14.24} & 98.02 & \textbf{17.20} & \textbf{80.19} & 30.55 & 80.23 & \textbf{37.48} & \textbf{92.58} & \textbf{13.78} \\
        \hline
		\bottomrule
		\end{tabular}}}
	\end{center}
\end{table*}

\subsection{Ablation Study}
We first performed an ablation study to figure out the influence of noise generation functions. We used DistilBERT for this study, because of its low model complexity, and all PLMs exhibited a similar trend in Table \ref{table2.Original_result}. The experiment was conducted for every possible combination of noise functions. For a fair comparision, we used the validation dataset to decide the best combination. The results are summarised in Table \ref{table3.AblationStudy}.  We observed that the best combination varies for each dataset. In detail, using the ``only Deletion'' function was the most efficient for the Corona Tweets. For the AG's News and Fake News dataset, the ``only Replacement'' function produced the best results. In addition, using only ``Deletion'' or ``Permutation'' has almost no impact on improving the OOD detection performance. We give further explanations regarding this phenomenon in Appendix~\ref{appendix.ablation}.

\subsection{Comparison with Fine-tuned Models}
\begin{figure}[t!]
	\centering
	\begin{subfigure}[b]{0.4\textwidth}
		\includegraphics[width=\linewidth]{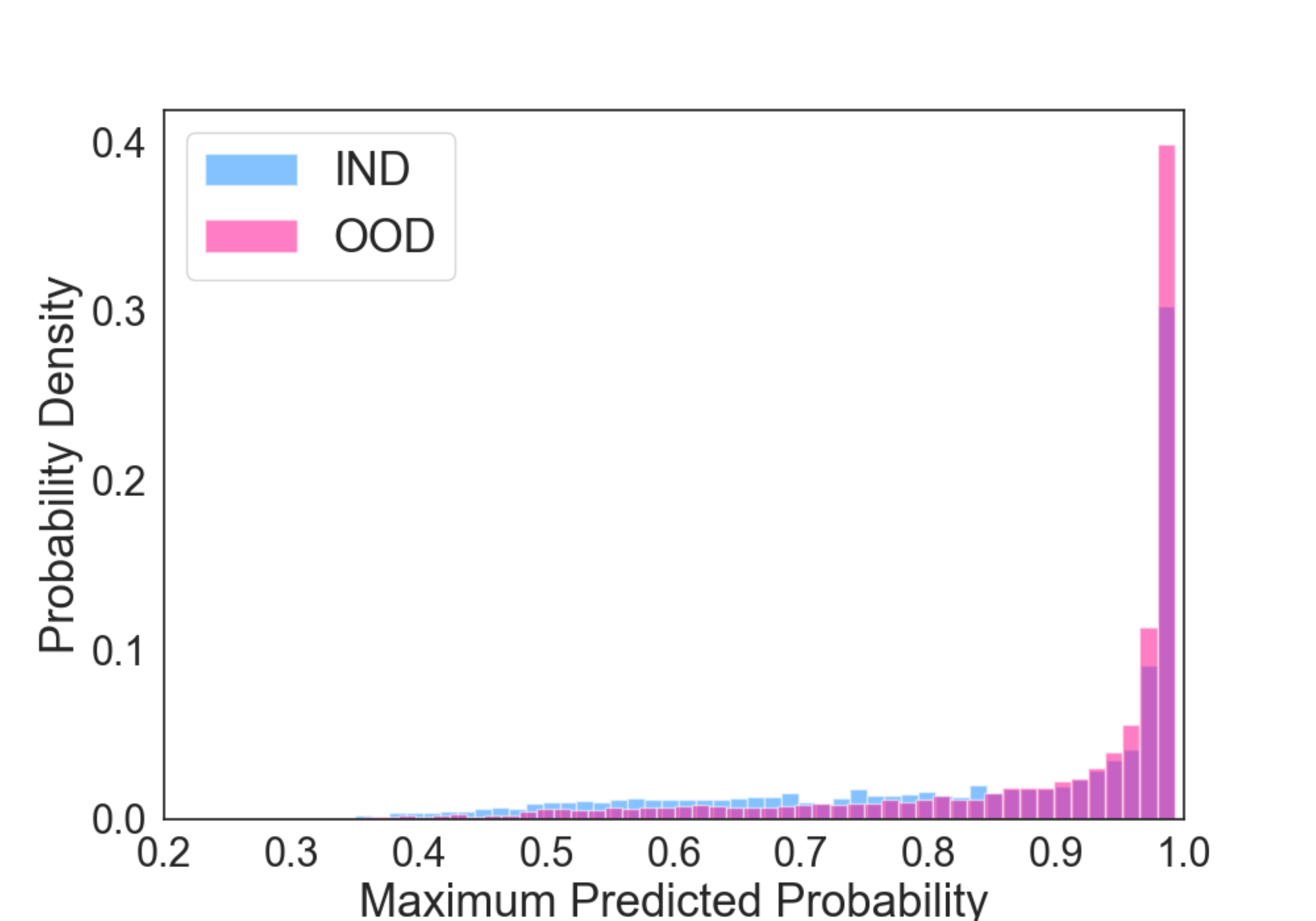}
		\caption{DistilBERT-base}
	\end{subfigure} \
	\begin{subfigure}[b]{0.4\textwidth}
		\includegraphics[width=\linewidth]{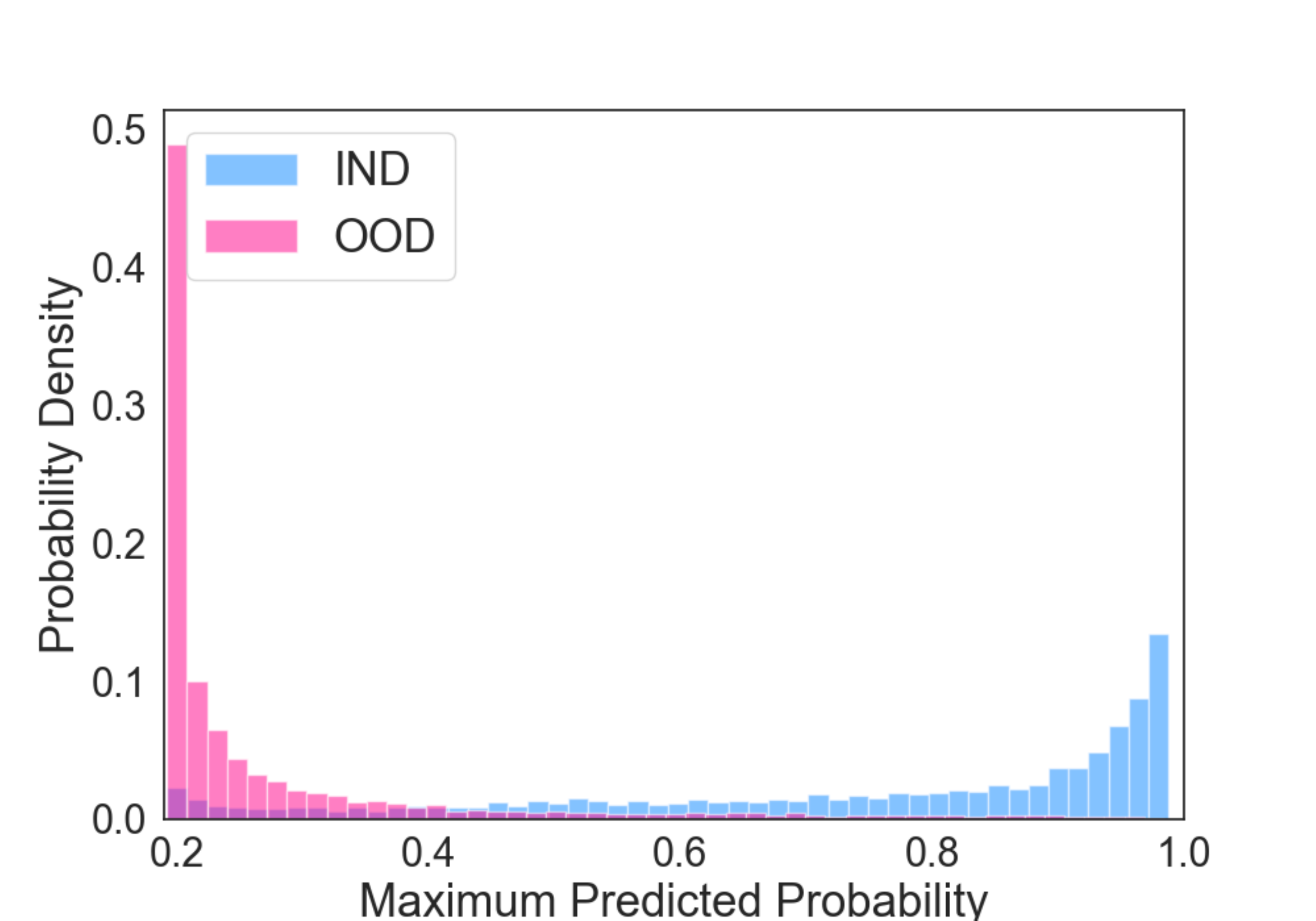}
		\caption{NoiER-DistilBERT-Best}
	\end{subfigure}%
	\caption{Maximum predicted probability distribution plots when the Corona Tweets dataset is IND.}
	\label{fig:ours_msp}
\end{figure}

The second experiment is comparing the performance of our proposed approach with that of the originally fine-tuned LMs. The results are summarised in Table \ref{table4.Original_compare}. We recorded the results of the ``Full'' model and the best noise function combination for each PLMs. Through the experiments, we confirmed that the ``Full'' model guarantees an improvement on the OOD detection ability regardless of the dataset and PLMs. Although the F1-score is marginally reduced by 1.1\% on average, the decrease is negligible, considering the promising improvement in OOD detection performance. Specifically, for the ``Full'' model case, the AUROC and EER values increased by 50\% and 36\% on average, respectively. For the ``Best'' case, the metrics were improved by 60\% and 49\% on average, respectively. The outperforming OOD detection ability of our proposed approach can also be verified by the distribution plots in Figure \ref{fig:ours_msp}. 

\subsection{Comparison with Baseline Models}
Finally, we conducted an experimental comparison with other baseline approaches. For a fair comparison, we selected OOD detectors that could be applied to PLMs while using  IND data only. We used DistilBERT \cite{DistilBERT} as a backbone pretrained model of this study because of its lower training time. The details of the baseline candidates are illustrated below.

\begin{enumerate}
    \item \textbf{MSP} \cite{hendrycks2017baseline}: We trained two variations of this approach, the original linear classifier (\textbf{MSP-Linear}) and the DistilBERT-based classifier (\textbf{MSP-DistilBERT}).
    
    \item \textbf{ODIN} \cite{liang2018enhancing}: We applied this method to MSP-Linear and MSP-DistilBERT, which are denoted \textbf{ODIN-Linear} and \textbf{ODIN-DistilBERT}, respectively. We used the same search space of \citet{liang2018enhancing} for selecting the hyperparameters.
    
    \item \textbf{AE} \cite{ryu2017neural}: Although the authors used a BiLSTM classifier, we replaced it with a BiLSTM-Self Attention model \cite{lin2017structured} (\textbf{AE-BiLSTMSA}) because of the original model's low performance on the Corona Tweets dataset. We also applied this method to DistilBERT (\textbf{AE-DistilBERT}).
    
    \item \textbf{ERPOG} \cite{zheng2020out}: We applied this method to DistilBERT, which is denoted \textbf{ERPOG-DistilBERT}.
    
    \item \textbf{Variants of our approach}: We additionally trained two variants of our approach. The first one is using IND sentences for minimising the ER term. This model can be considered as a special case of generalised entropy regularisation \cite{GER}. The second model adds continuous Gaussian noise to the distributed sentence representations \citet{zhang2018word}. We denote the aforementioned baselines  \textbf{GER-DistilBERT} and \textbf{CNER-DistilBERT}, respectively.
\end{enumerate}

Table \ref{table5.Benchmark_test} describes the comparison results. The IOD value of \textbf{AE} models are omitted, because they detect OOD by using reconstruction errors. The results reveal that the proposed approach better detects OOD compared to the baselines. The results show that our approach is capable of detecting OOD instances well for all datasets. For the AG's News dataset, \textbf{AE-BiLSTMSA} and \textbf{ERPOG-DistilBERT} produced promising results among the baselines. They exhibited comparable but slightly better results compared to the ``Full'' model of our approach in terms of OOD detection ability. However, all the baseline models fell short of distinguishing OOD instances in the Fake News and Corona Tweets dataset. In particular, \textbf{ERPOG-DistilBERT} entirely failed to detect OOD in both datasets. We suspect that this result stems from the relatively longer sentence length compared to the AG's News dataset. The performance of the LSTM-based Seq2Seq model dwindles as the length of the sequence goes longer \cite{bahdanau2014neural}, which leads the pseudo-OOD generator to produce defective sentences. Therefore, the ERPOG model is trained with corrupted sentences, which inevitably degrades both the classification and the OOD detection performance. This result reveals the drawback of OOD generator-based approaches.

It is also worth mentioning that the variants of our approach have no impact on the OOD detection performance. Notably, generating continuous noise samples aggravated the distribution collapse issue. This may be due to the natural structure of text data as a discrete token sequence, which makes them suffer from latent vacancy problem \cite{xu2020variational}. Likewise, the experimental results demonstrate our approach's effectiveness, which leverages discrete pseudo OOD examples generated in a model-free~way. 

\section{Summary and Outlook}
In this paper, we have shown that the distribution collapse is a prevalent issue that frequently occurs in PLMs. This is a problematic issue, because it degrades the reliability of the models. To solve this issue, we propose a simple but effective approach: noise entropy regularisation. The proposed method generates noise sentences, which share a similar syntactic form with IND sentences, and leverages them to minimise entropy regularisation loss. Extensive experiments reveal that the proposed approach significantly improves the OOD detection ability of the original PLMs, while preserving their high performance on classification. Also, it outperforms baseline models designed for OOD text detection. Moreover, our approach is advantageous in terms of its practicality, because additional data or auxiliary models are unnecessary.

As for future works, an interesting topic is to expand our approach to sequential labeling tasks, such as named entity recognition (NER). Also, our approach has a drawback in that it  relies on the randomised process, which could generate different outcomes for each training process. Therefore, implementing a more stable method is another interesting research direction.

\bibliography{acl2021}
\bibliographystyle{acl_natbib}

\clearpage
\appendix
\addcontentsline{toc}{section}{Appendix}
\section{Supplemental Material}
\subsection{Evaluation Metrics} \label{appendix.eval_metric}
In addition to the IOD value, we leveraged two widely used metrics: AUROC and EER for measuring OOD detection performance. AUROC is the area under the receiver operating characteristic curve, which is created by plotting the true positive rate (TPR) against the false positive rate (FPR) at diverse levels of thresholds \cite{chen2005application}. EER is the intersection point where the false rejection rate (FRR) and the false acceptance rate (FAR) meet \cite{9399070}. Therefore, lower EER values imply that a model is more accurate.

\subsection{Hyperparameter Search} \label{appendix.hps}
\begin{figure*}[t!]
	\centering
	\begin{subfigure}[b]{0.32\textwidth}
		\includegraphics[width=\linewidth]{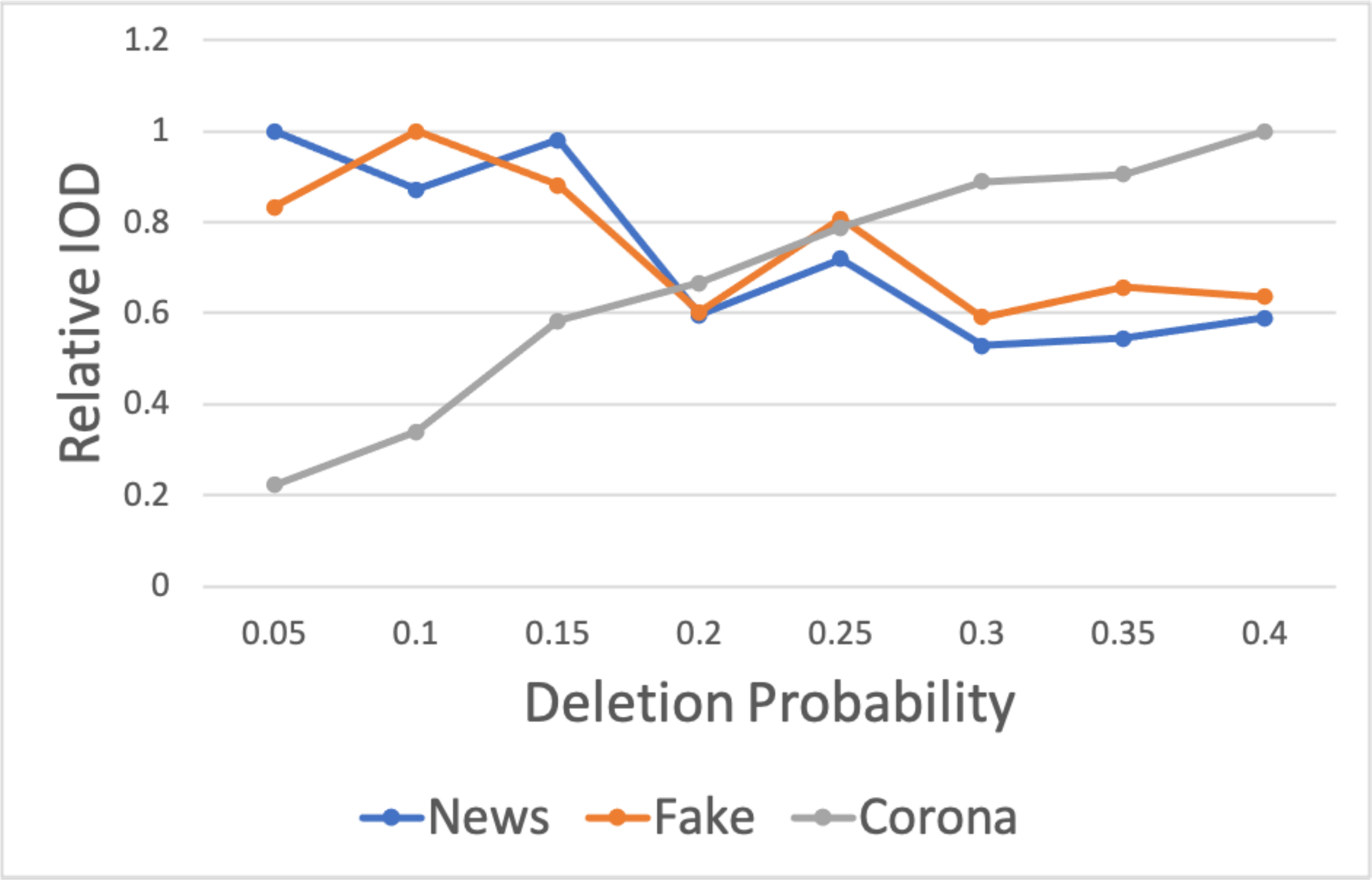}
		\caption{Deletion ($p_{del}$)}
	\end{subfigure} 
	\begin{subfigure}[b]{0.32\textwidth}
		\includegraphics[width=\linewidth]{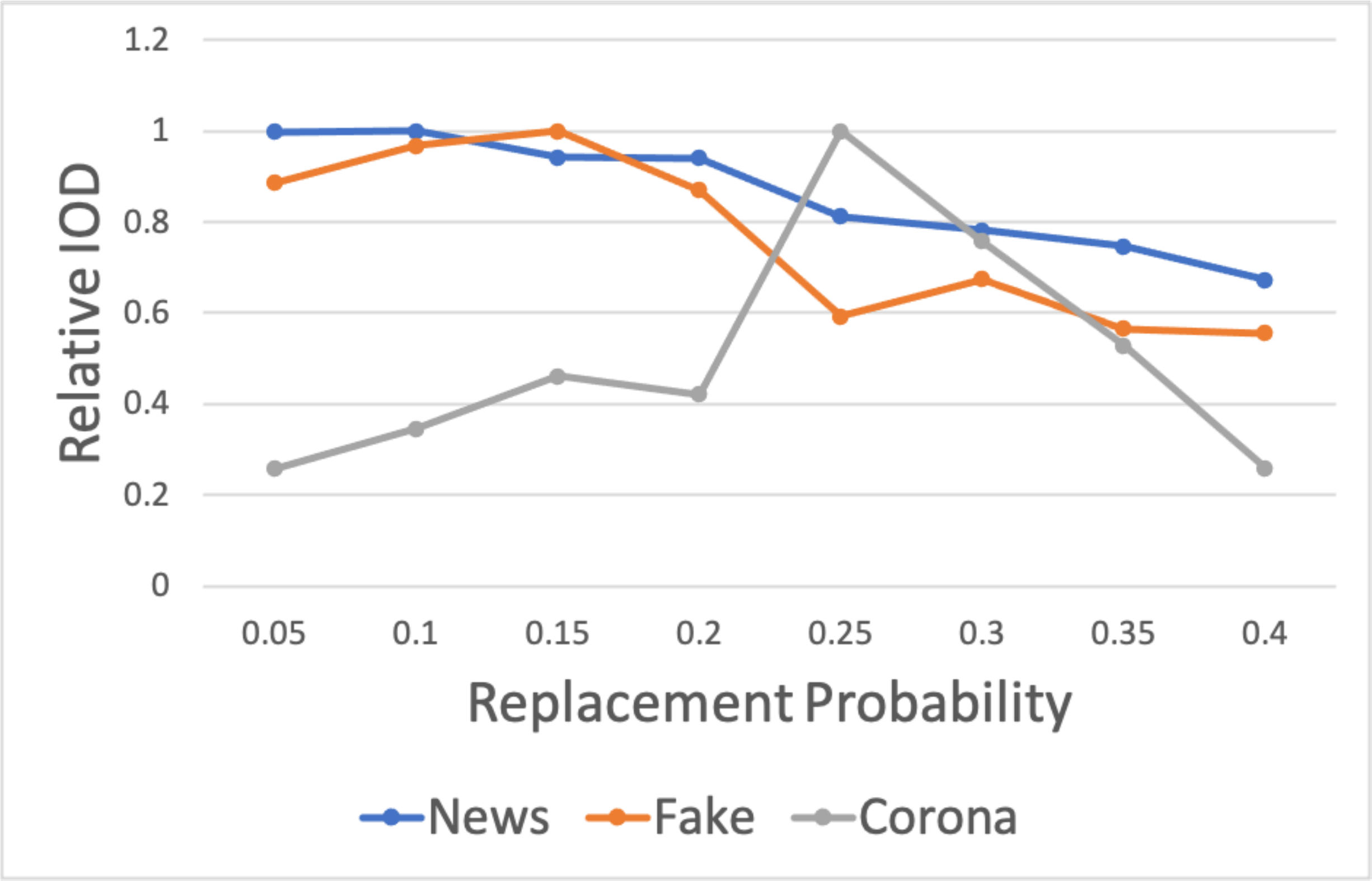}
		\caption{Replacement ($p_{repl}$)}
	\end{subfigure}%
	\begin{subfigure}[b]{0.32\textwidth}
		\includegraphics[width=\linewidth]{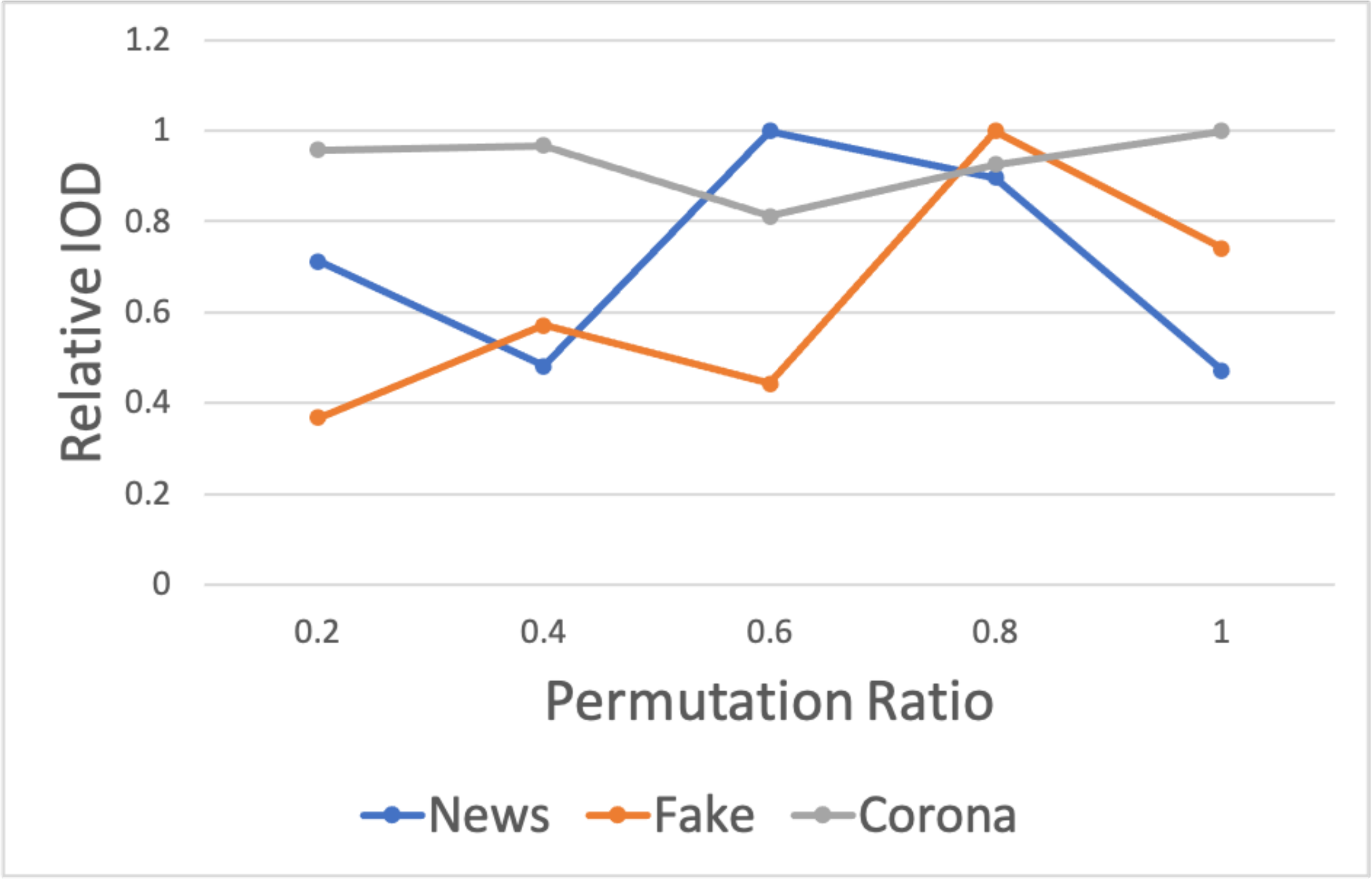}
		\caption{Permutation ($r_{perm}$)}
	\end{subfigure}%
	\caption{Plots for the relative average IOD values for deciding hyper-parameters of noise generation functions.}
	\label{fig:hps_plot}
\end{figure*}

In this section, we present more detailed explanations regarding the noise function's hyperparameter search. We have three hyperparameters to decide: $p_{del}$, $p_{repl}$, and $r_{perm}$ for deletion, replacement, and permutation, respectively. For the probability $p_{del}$ and $p_{repl}$, we choose from evenly spaced 8 numbers starting from 0.05 and ending at 0.4. For the ratio $r_{perm}$, we select among 0.2, 0.4, 0.6, 0.8, and 1.0. To decide the optimum figures, we leveraged the IOD value of the validation dataset, which is independent from the test dataset. We repeated the training 10 times for each hyperparameter setting and leveraged DistilBERT as a backbone model. Figure \ref{fig:hps_plot} presents the relative average IOD values on the hyperparameter settings for each noise functions. The selected values are summarised in Table \ref{table.hps}. We found that there exists a strong correlation between the average length of the sentences included in the dataset and the noise generation hyperparameter values. We conjecture that this is because the high noise parameter values on short sentences significantly change the IND sentences, violating Condition 1  in Section \ref{noier}. On the contrary, low noise parameter values on long sentences make IND and perturbed sentences indistinguishable, which violates Condition 2.

\subsection{Correlation between IOD and AUROC}
A text classification model that possesses a good OOD detection ability should produce high IOD and AUROC scores. However, this does not imply that the IOD value has a positive correlation with the AUROC score. The model that produces a high IOD value also generates a high AUROC score. However, the vice versa case does not hold in a certain situation. This is because the IOD value merely measures the distribution disparity, whereas AUROC is an evaluation metric that belongs to the accuracy category.

\begin{table}[t!]
	\begin{center}
		\caption{Summary of the selected hyper-parameters} \label{table.hps}%
		\renewcommand{\arraystretch}{1.3}
		\footnotesize{
			\centering{\setlength\tabcolsep{3pt}
				\begin{tabular}{cccc}
					\toprule
					\hline
					& AG's News & Fake News & Corona Tweets \\
					\hline
					$p_{del}$ & 0.05 & 0.1 & 0.4 \\
				    $p_{repl}$ & 0.1 & 0.15 & 0.25 \\
					$r_{perm}$ & 0.6 & 0.8 & 1.0 \\
					\hline
					\bottomrule	
		\end{tabular}}}
	\end{center}
\end{table}

\begin{figure}[t!]
	\centering
	\begin{subfigure}[b]{0.4\textwidth}
		\includegraphics[width=\linewidth]{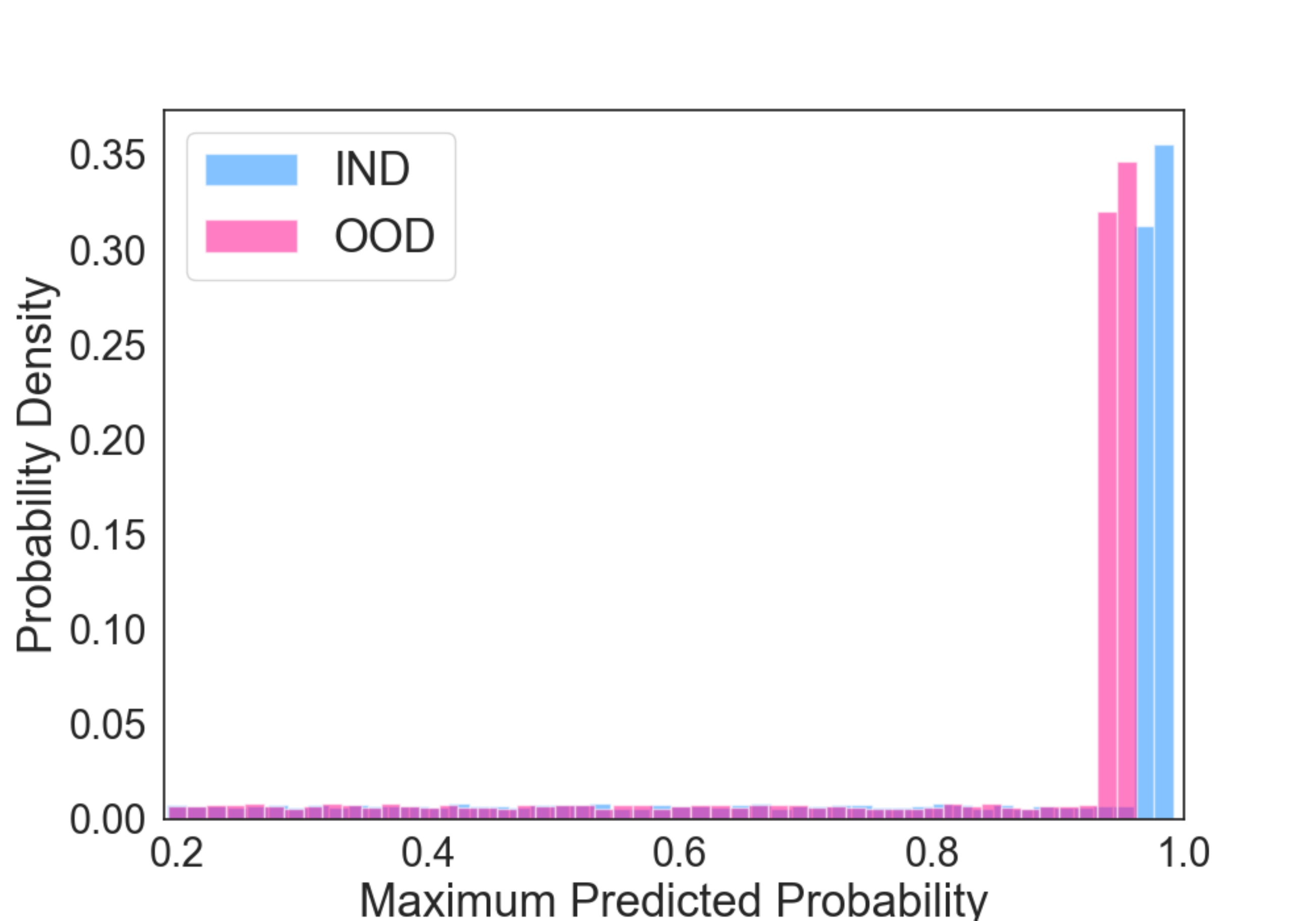}
		\caption{low IOD but high AUROC}\label{case1}
	\end{subfigure}%
	\caption{Maximum predicted probability distribution plot for explaining uncorrelatedness between IOD and AUROC.}
	\label{fig:iod_auroc_example}
\end{figure}

\begin{figure*}[t!]
	\centering
	\begin{subfigure}[b]{0.32\textwidth}
		\includegraphics[width=\linewidth]{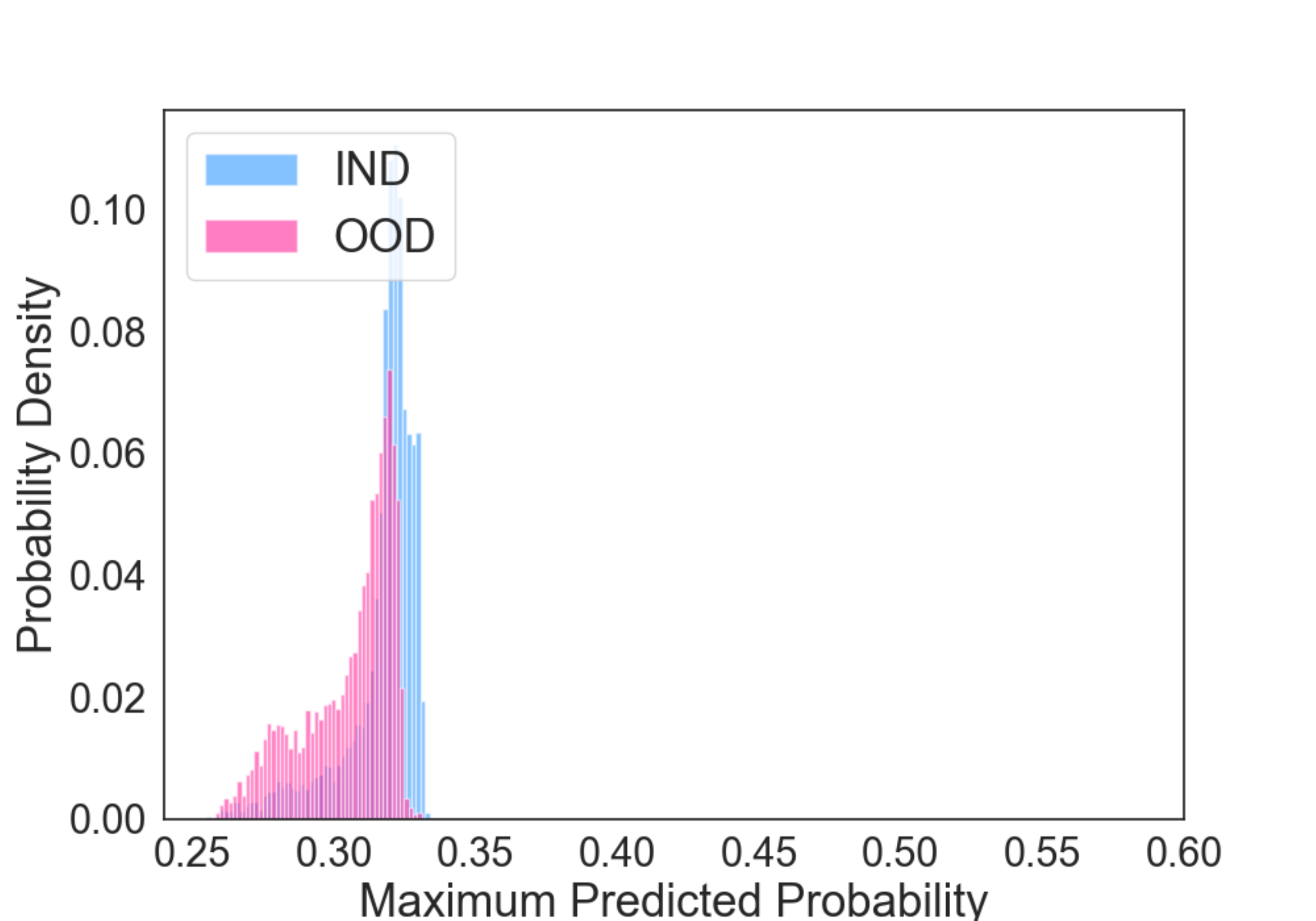}
		\caption{AG's News}
	\end{subfigure} 
	\begin{subfigure}[b]{0.32\textwidth}
		\includegraphics[width=\linewidth]{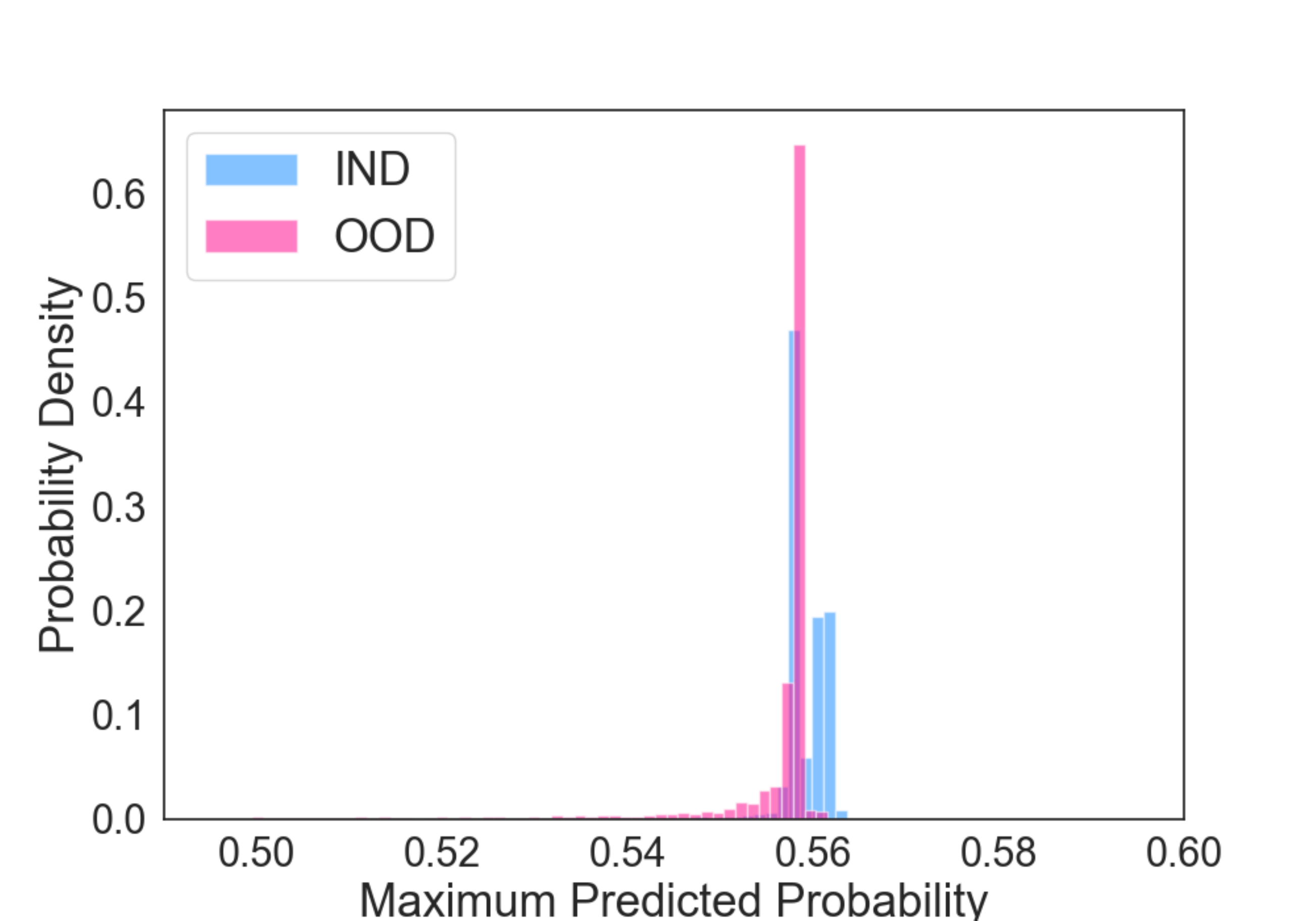}
		\caption{Fake News}
	\end{subfigure}%
	\begin{subfigure}[b]{0.32\textwidth}
		\includegraphics[width=\linewidth]{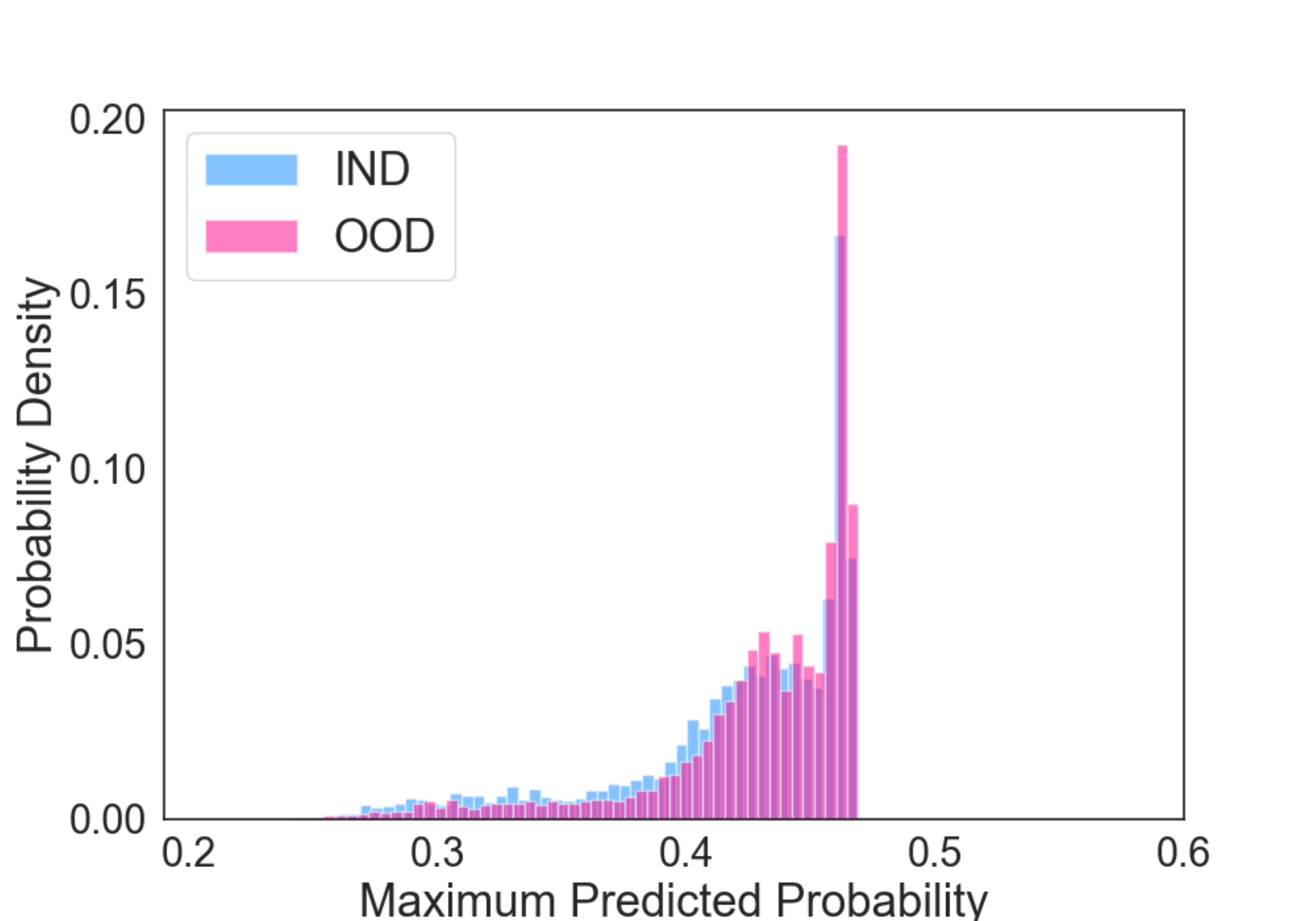}
		\caption{Corona Tweets}
	\end{subfigure}%
	\caption{Maximum predicted probability distribution plots of GER-DistilBERT.}
	\label{fig:ger_plot}
\end{figure*}


\begin{algorithm}
\caption{Generating noise sentences.}
	\textbf{Input:} An input sentence $s=(w_1, w_2, ..., w_n)$, where $w_i$ is an $i^{th}$ word of the sentence $s$. The hyperparameters: $p_{del}$, $p_{repl}$, and $r_{perm}$.\\
	\textbf{Output:} A noise-added sentence 
	\begin{algorithmic}[1]
		\State $\hat{s}$ $\leftarrow$ \textrm{COPY}($s$)\Comment{Copy the input}
		\While{$s == \hat{s}$}
		\State $p$ $\leftarrow$ \textrm{UNIFORM(0,1)}
		\If{$p$ $\leq \frac{1}{3}$ } \Comment{Delete words}
		\State s $\leftarrow$ \textrm{DELETEWORD(s, $p_{del}$)}
		\ElsIf{$p$ $\geq \frac{2}{3}$} \Comment{Permute words}
		\State L $\leftarrow$ len(s)
		\State s $\leftarrow$ \textrm{PERMUTEWORD(s, $r_{perm}$)}
		\Else \Comment{Replace words}
		\State s $\leftarrow$ \textrm{REPLACEWORD(s, $p_{repl}$)}
		\EndIf
		\EndWhile
	\end{algorithmic} 
	\label{algo.noise}
\end{algorithm} 
\subsubsection{Low IOD but High AUROC}
Figure \ref{case1} well illustrates the corresponding situation. In this example, the AUROC score is quite high, because we can distinguish IND and OOD examples, provided we set a threshold to about 0.95. However, the IOD value is close to 0, because the two distributions are almost similar. In our experiments, the results of \textbf{GER-DistilBERT} on the AG's News and Fake News dataset, which are provided in Table \ref{table5.Benchmark_test} and Figure \ref{fig:ger_plot}, correspond to this case.

Although a model can generate a high AUROC score but a low IOD value, it is obvious that a model that produces both a high IOD value and a high AUROC score is a better classification model, because it is much easier to set a threshold for rejecting OOD. Therefore, both metrics have to be considered to fully evaluate the model's OOD detection performance.

\subsection{Noise Generation Algorithm}
The overall process for generating noise sentences is described in Algorithm \ref{algo.noise}. It randomly decides one function from the three candidates. For the ablation study, we manipulated the random probability $p$ to prevent certain functions from being selected.

\begin{table*}[t!]
	\begin{center}
		\caption{Generated noise examples for the AG's News dataset} \label{table.noise_examples_AG}%
		\renewcommand{\arraystretch}{1.5}
		\footnotesize{
			\centering{\setlength\tabcolsep{2.5pt}
		\begin{tabular}{l|c|c|l}
		\toprule
		\hline
		\makecell[c]{Original sentence} & Label & \multicolumn{2}{c}{Generated noise }\\ \hline
		
		\multirow{3}{*}{\makecell{HP shares tumble on profit news}} & \multirow{3}{*}{\makecell{Business}}
        & Del & HP shares tumble profit news \\
        & & Permute & HP tumble shares on news profit \\
        & & Repl & 7Z shares tumble GE profit news \\ \hline
        
        \multirow{3}{*}{\makecell{How mars fooled the world}} & \multirow{3}{*}{\makecell{Sci/Tech}}
        & Del & How mars fooled world \\
        & & Permute & mars How fooled the world \\
        & & Repl & 06B mars fooled the SD8EG \\ \hline

        \multirow{3}{*}{\makecell{Venezuela opposition holds recall vote}} & \multirow{3}{*}{\makecell{World}}
        & Del & Venezuela opposition recall vote \\
        & & Permute & opposition Venezuela holds recall vote \\
        & & Repl & Venezuela opposition holds 0MUFHF vote \\ \hline        
        
        \multirow{3}{*}{\makecell[l]{Basketball: China \#39;s Yao takes anger \\ out on New Zealand}} & \multirow{3}{*}{\makecell{Sports}}
        & Del & Basketball: china \#39;s Yao takes out on New Zealand \\
        & & Permute & China basketball: \#39;s anger out Yao takes New on Zealand \\
        & & Repl & Basketball: HRTE5 \#39;s Yao takes anger out on New FZH611A \\ \hline
        
	    \bottomrule
		\end{tabular}}}
	\end{center}
\end{table*}

\subsection{Ablation Study Result Analysis}\label{appendix.ablation}
In this section, we describe a more detailed explanation of the ablation study results. The notable characteristic is that the ``Deletion'' and ``Permutation'' functions are unprofitable, or even worse, work negatively on the AG's News and Fake News dataset. We conjecture that this phenomenon is due to the length of the input sentence. As illustrated in Appendix \ref{appendix.hps}, the shorter the input sentence is, the fewer noise is added to prevent violating Condition 1 in Section \ref{noier}. However, unlike the ``Replacement'' function, which introduces external words into the sentence, ``Deletion'' and ``Permutation'' functions only reorganise words within a sentence. These characteristics inevitably increase the possibility of the sentence being considered as IND. As a result, Condition 2,  in Section \ref{noier} is violated, which causes a negative influence on the model's OOD detection ability. Table \ref{table.noise_examples_AG} contains the examples of generated noise sentences on the AG's News dataset. Unlike the ``Replacement'' function, considering the result of the ``Deletion'' and ``Permutation'' functions as OOD sentences is not reasonable.

The experimental results presented in Table \ref{table3.AblationStudy} also support the relationship between the input sentence's length and the effect of the ``Deletion'' and ``Permutation'' functions. For the AG's News dataset, both the ``only Deletion'' and the ``only Permutation'' model significantly performed less than the original fine-tuned model under 0.01 confidence level. However, when it comes to the Fake News dataset, the ``only Deletion'' model's results showed no difference with the original fine-tuned model. Although the ``only Permutation'' model did not perform well again, it produced a higher $p$-value. Finally, for the Corona Tweets dataset, both the ``only Deletion'' and the ``only Permutation'' model outperformed the original fine-tuned model under 0.01 confidence level. Therefore, we recommend users to remove the ``Deletion'' and ``Permutation'' functions, if the dataset that they use to apply our approach on contains many short sentences.

\subsection{Example Graphs}
\begin{figure*}[t!]
	\begin{subfigure}[b]{0.32\textwidth}
		\includegraphics[width=\linewidth]{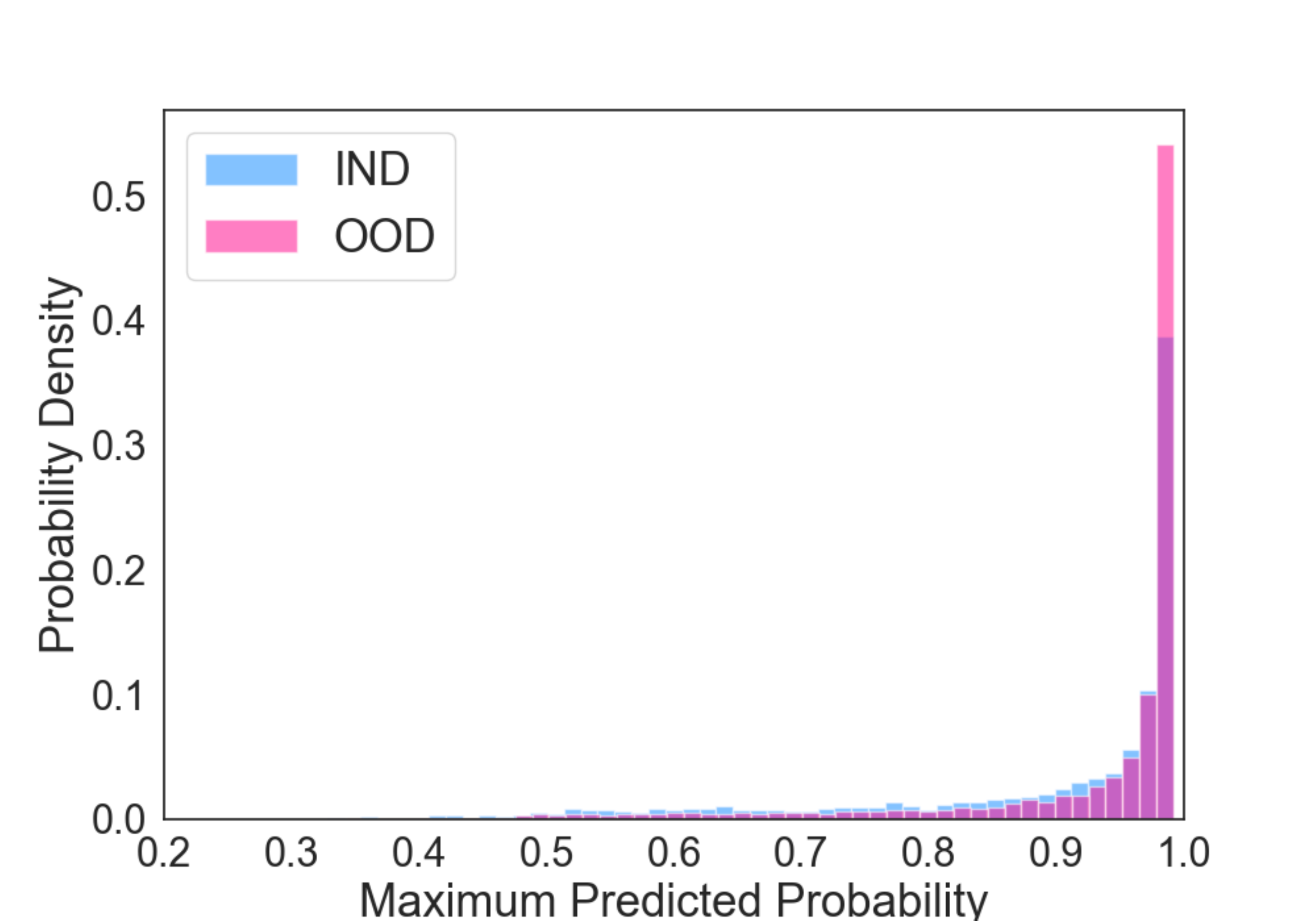}
		\caption{BERT-base}
	\end{subfigure}
	\begin{subfigure}[b]{0.32\textwidth}
		\includegraphics[width=\linewidth]{images/original/distilbert_base_corona_dist.pdf}
		\caption{DistilBERT-base}
	\end{subfigure}
	\begin{subfigure}[b]{0.32\textwidth}
		\includegraphics[width=\linewidth]{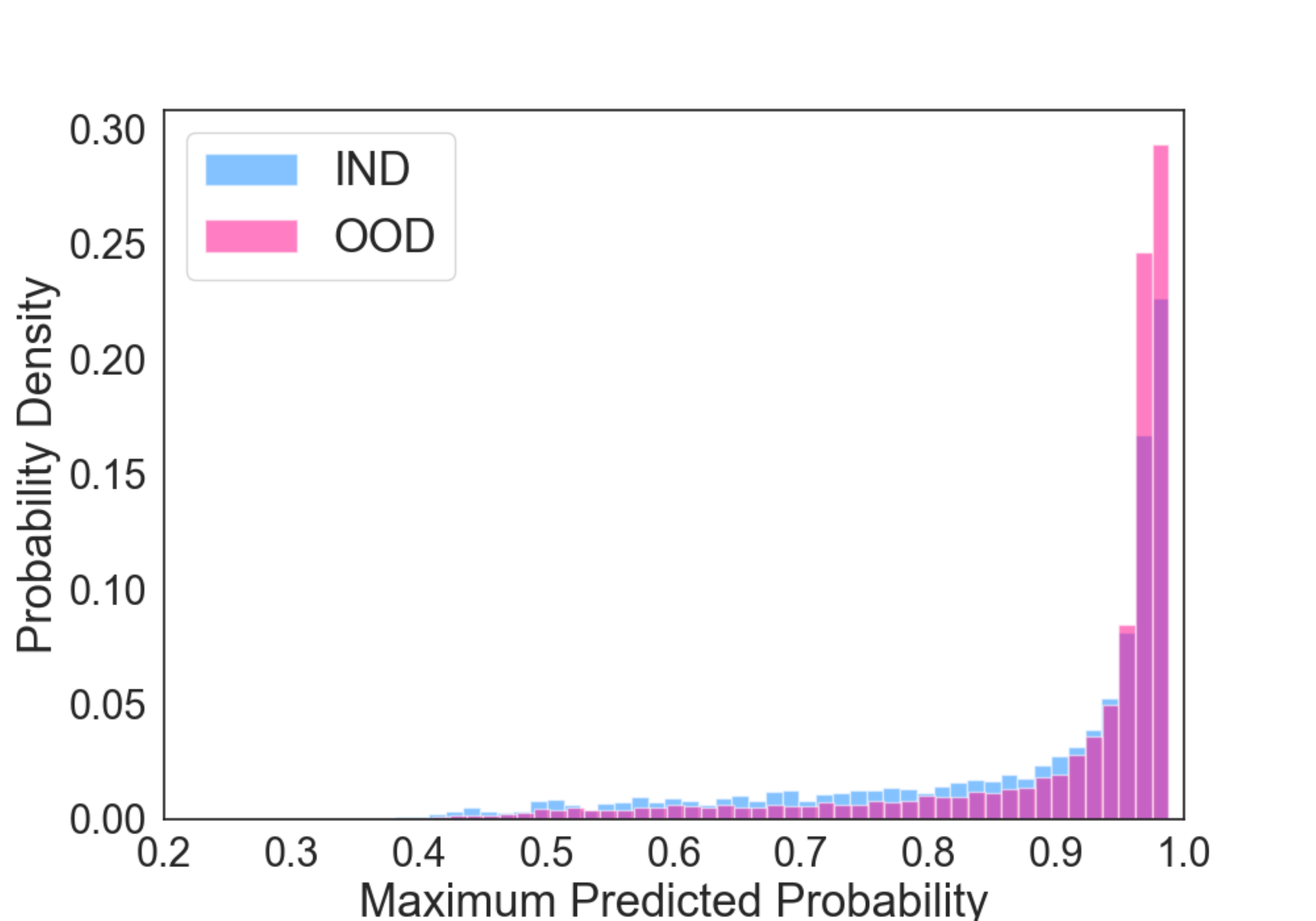}
		\caption{ELECTRA}
	\end{subfigure}
	
	\begin{subfigure}[b]{0.32\textwidth}
		\includegraphics[width=\linewidth]{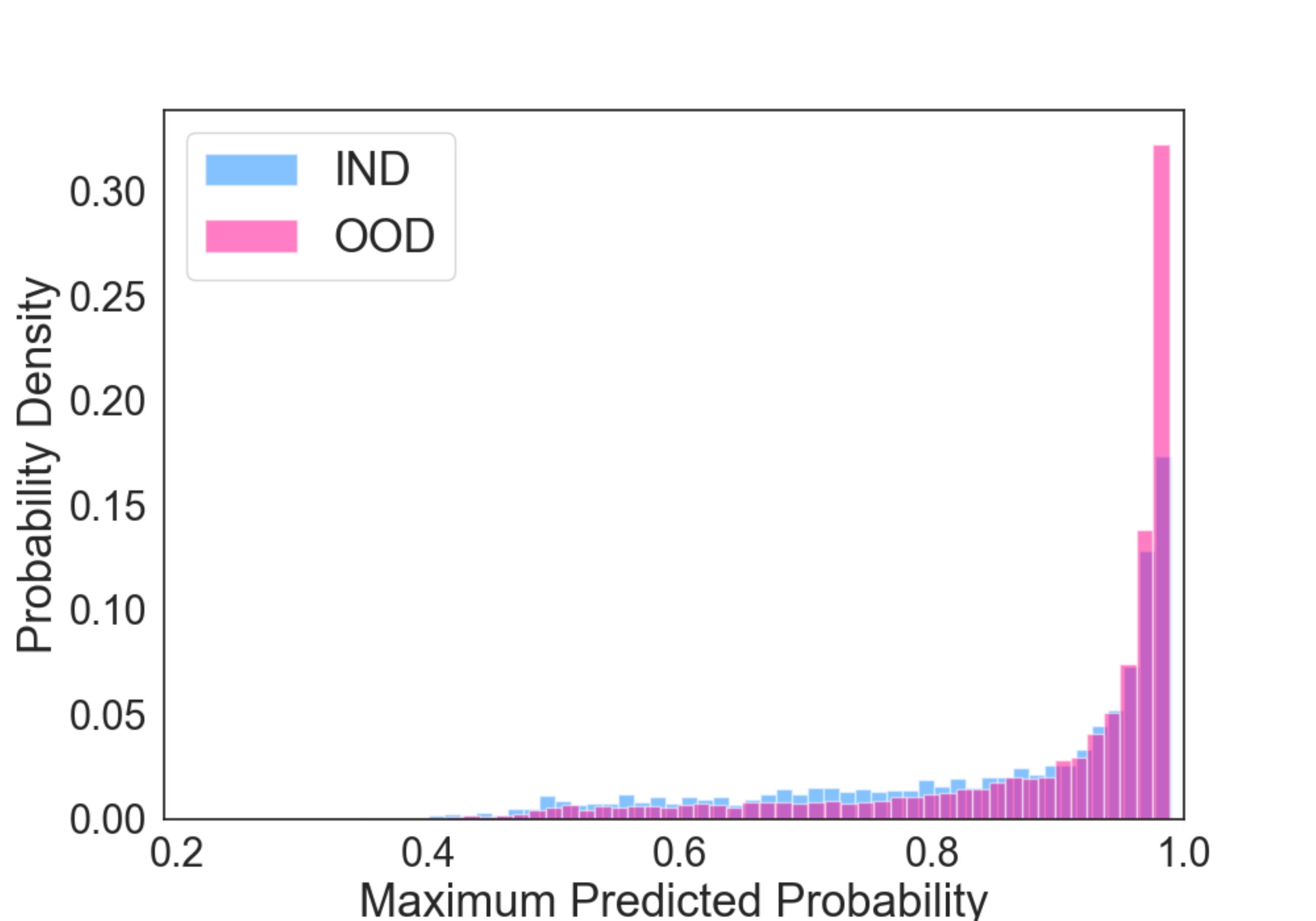}
		\caption{SqueezeBERT}
	\end{subfigure}
	\begin{subfigure}[b]{0.32\textwidth}
		\includegraphics[width=\linewidth]{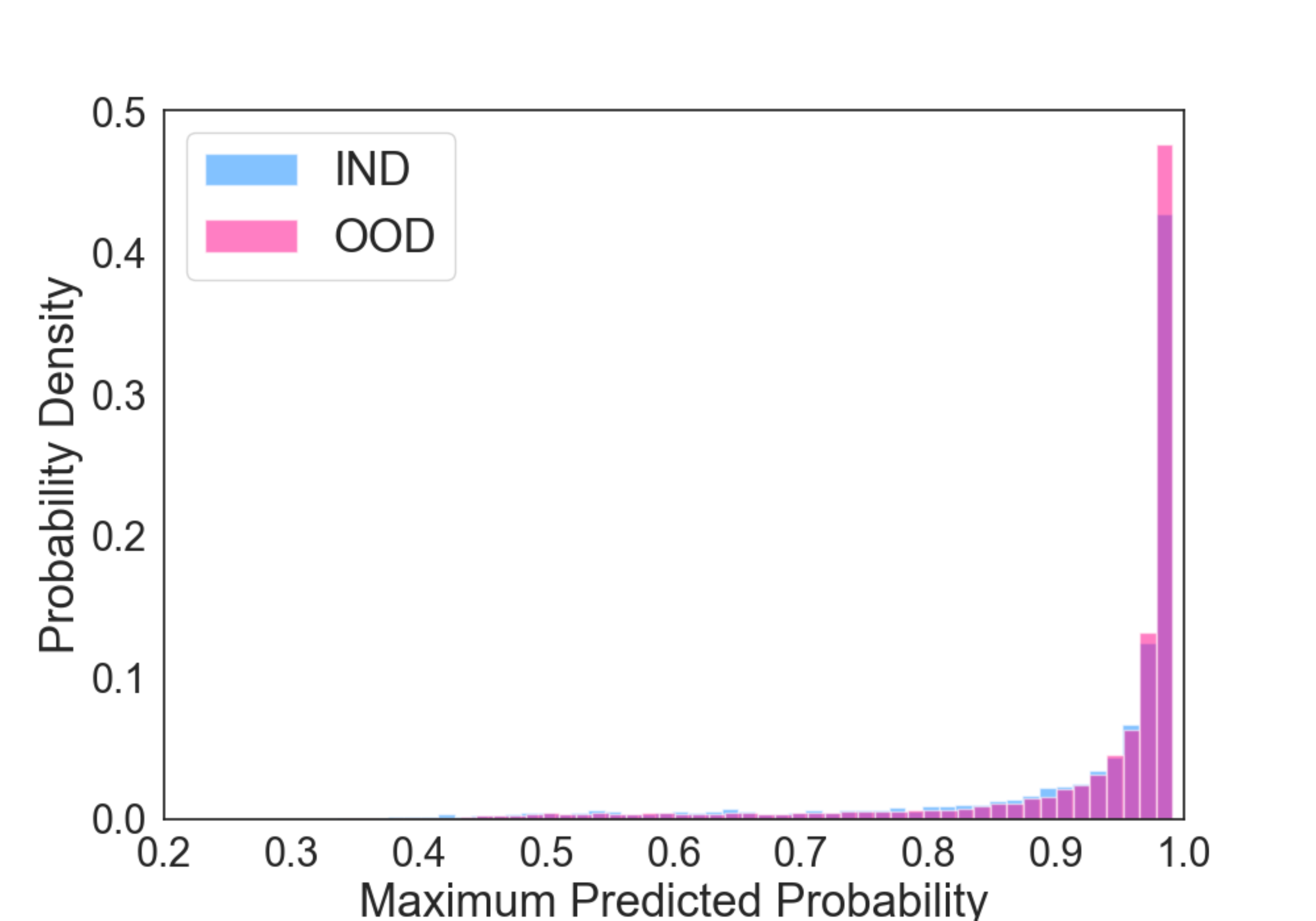}
		\caption{ERNIE 2.0}
	\end{subfigure}
	
	\caption{Maximum predicted probability distribution plots of original fine-tuned models when Corona Tweets data are IND.}
	\label{fig:ours_corona_original}
\end{figure*}

\begin{figure*}[t!]
	\begin{subfigure}[b]{0.32\textwidth}
		\includegraphics[width=\linewidth]{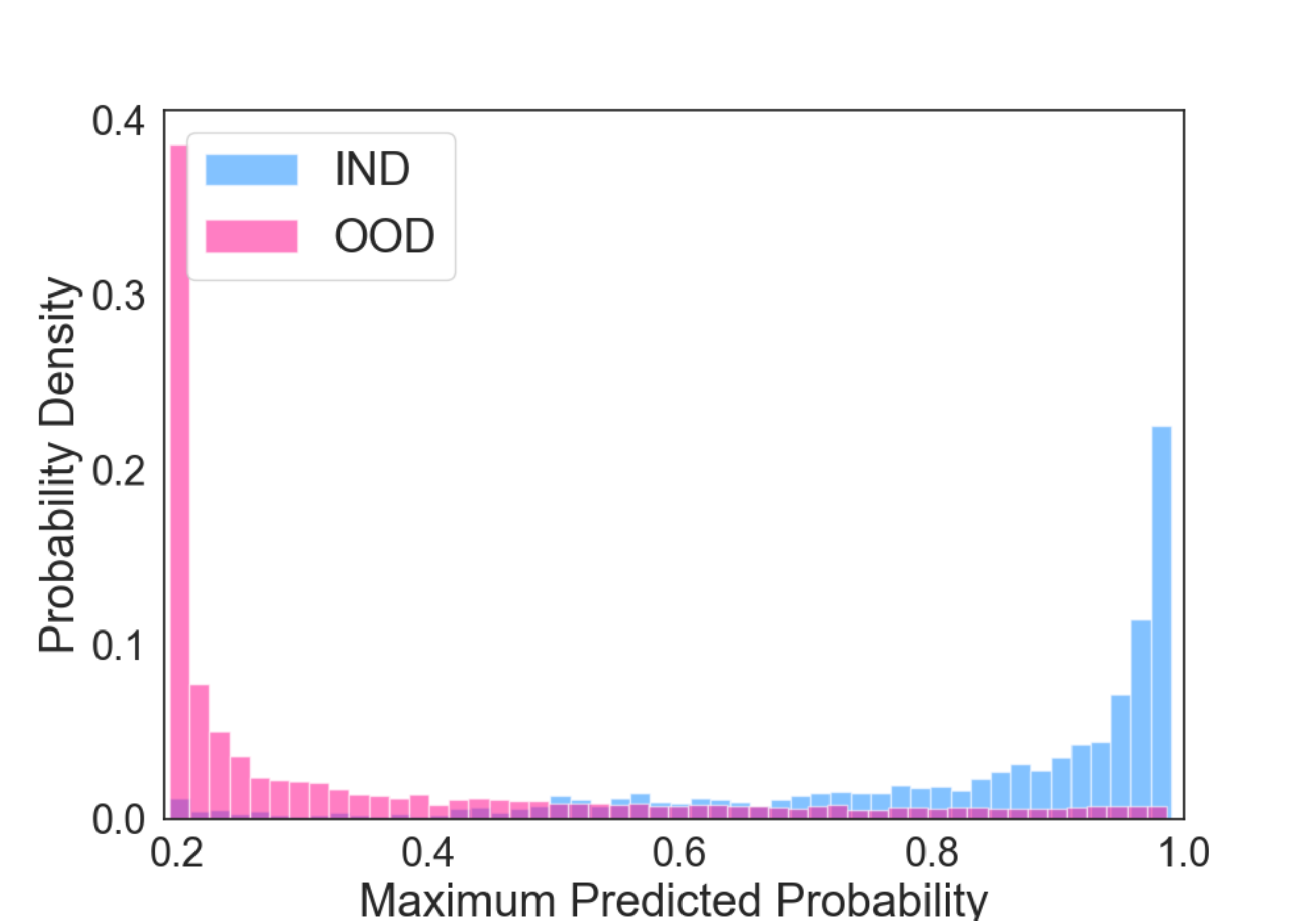}
		\caption{BERT-base}
	\end{subfigure}
	\begin{subfigure}[b]{0.32\textwidth}
		\includegraphics[width=\linewidth]{images/ours/distilbert_base_corona_best_dist.pdf}
		\caption{DistilBERT-base}
	\end{subfigure}
	\begin{subfigure}[b]{0.32\textwidth}
		\includegraphics[width=\linewidth]{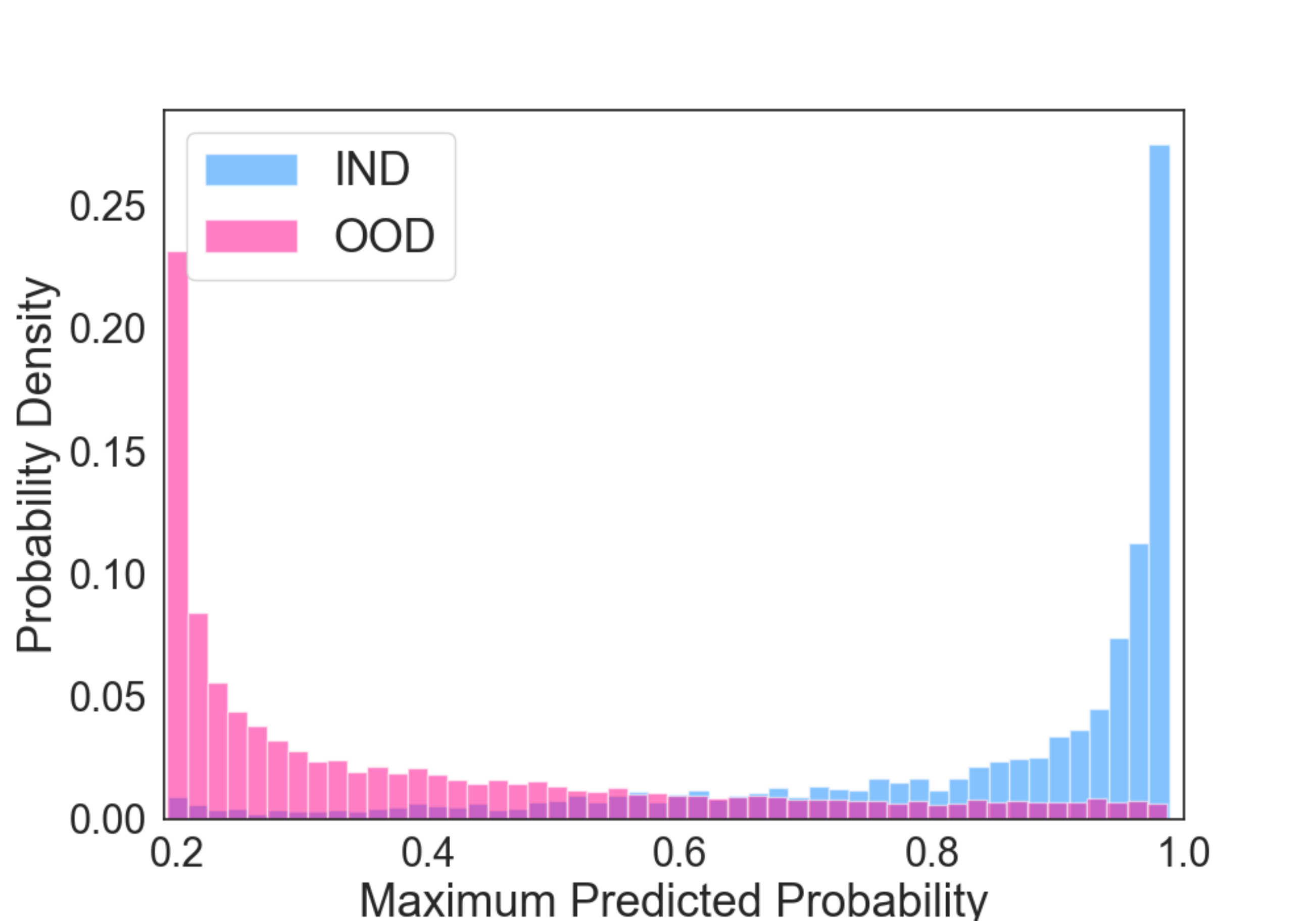}
		\caption{ELECTRA}
	\end{subfigure}
	
	\begin{subfigure}[b]{0.32\textwidth}
		\includegraphics[width=\linewidth]{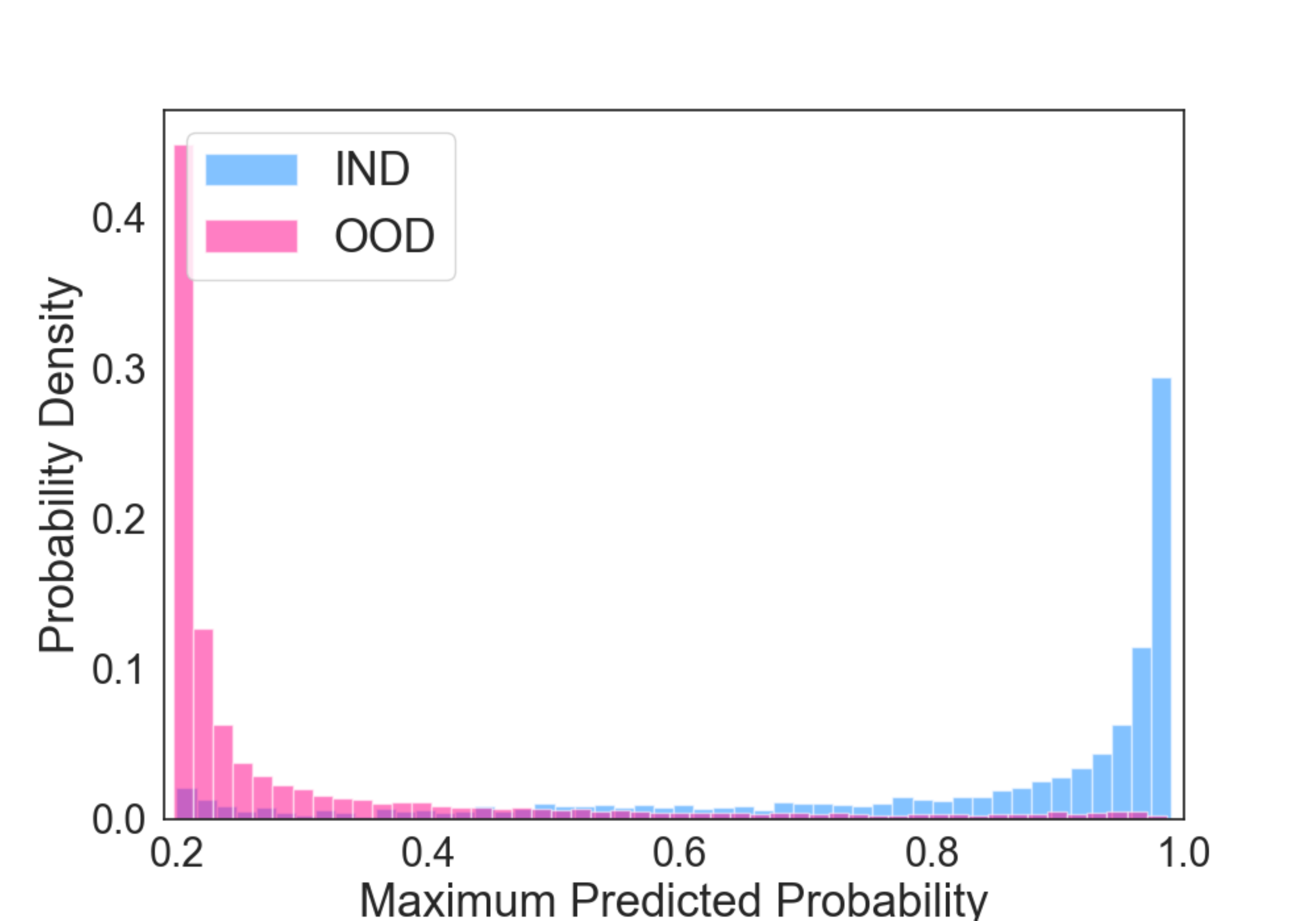}
		\caption{SqueezeBERT}
	\end{subfigure}
	\begin{subfigure}[b]{0.32\textwidth}
		\includegraphics[width=\linewidth]{images/ours/electra_corona_best_dist.pdf}
		\caption{ERNIE 2.0}
	\end{subfigure}
	
	\caption{Maximum predicted probability distribution plots of NoiER-BEST models when Corona Tweets data are IND.}
	\label{fig:ours_corona_best}
\end{figure*}

We added examples of the maximum predicted probability distribution plots of original fine-tuned models and those of our proposed approach. Figures \ref{fig:ours_corona_original} and  \ref{fig:ours_corona_best} present the distribution plots of the fine-tuned models and the NoiER approach on the Corona Tweets dataset. It can be easily seen that the proposed approach distinguishes OOD sentences much better than the original fine-tuned models.

\end{document}